\documentclass{article}

\usepackage[final,nonatbib]{neurips_2019}

\usepackage[utf8]{inputenc} 
\usepackage[T1]{fontenc}    
\usepackage{hyperref}       
\usepackage{url}            
\usepackage{booktabs}       
\usepackage{amsfonts}       
\usepackage{nicefrac}       
\usepackage{microtype}      

\usepackage{float}
\usepackage{subfig}
\usepackage{graphicx}
\usepackage{multirow}
\usepackage{framed}
\usepackage{amsmath}

\usepackage[numbers]{natbib}

\title{On Learning Paradigms for the\\ Travelling Salesman Problem}

\author{%
  Chaitanya K. Joshi$^{1}$, 
  Thomas Laurent$^{2}$, 
  and Xavier Bresson$^{1}$ \\
  $^{1}$School of Computer Science and Engineering, 
  Nanyang Technological University, Singapore \\
  $^{2}$Department of Mathematics, 
  Loyola Marymount University \\
  \texttt{\{chaitanya.joshi, xbresson\}@ntu.edu.sg, tlaurent@lmu.edu} \\
}

\begin{document}

\maketitle

\begin{abstract}
We explore the impact of learning paradigms on training deep neural networks for the Travelling Salesman Problem. 
We design controlled experiments to train supervised learning (SL) and reinforcement learning (RL) models on fixed graph sizes up to $100$ nodes, and evaluate them on variable sized graphs up to $500$ nodes.
Beyond not needing labelled data, our results reveal favorable properties of RL over SL:
RL training leads to better \textit{emergent} generalization to variable graph sizes and 
is a key component for learning scale-invariant solvers for novel combinatorial problems.
\footnote{Code available: \url{https://github.com/chaitjo/learning-paradigms-for-tsp}}
\end{abstract}

\section{Introduction}

The Travelling Salesman Problem (TSP)
is one of the most intensively studied combinatorial optimization problems in the Operations Research community
and is the backbone of industries such as transportation, logistics and scheduling.
Being an NP-hard graph problem, finding optimal TSP solutions is intractable at large scales above thousands of nodes. 
In practice, the Concorde TSP solver \citep{applegate2006traveling} uses carefully handcrafted heuristics to find approximate solutions up to tens of thousands of nodes.
Unfortunately, powerful OR solvers such as Concorde are problem-specific; their development for new problems requires significant time and specialized knowledge.

An alternate approach by the Machine Learning community is to develop generic learning algorithms which can be trained to solve \textit{any} combinatorial problem from problem instances themselves \citep{smith1999neural,bengio2018machine}.
Using 2D Euclidean TSP as a representative of practical combinatorial problems,
recent learning-based approaches \citep{khalil2017learning,deudon2018learning,kool2018attention} have leveraged advances in graph representation learning \citep{bruna2014spectral,defferrard2016convolutional,sukhbaatar2016gcn,kipf2017semi,hamilton2017inductive,monti2017geometric} to operate directly on the problem's graph structure and perform competitively with Concorde on problem instances of fixed/trivially small graphs.

A key design choice for scaling these approaches to real-world problem sizes is the learning paradigm: supervised learning (SL) or reinforcement learning (RL). 
As noted in \cite{bengio2018machine}, the performance of SL-based models depends on the availability of large sets of optimal or high-quality instance-solution pairs.
Although RL is known to be less sample efficient than SL, it does not require labelled instances.
As long as a problem can be formulated via a reward signal for making sequential decisions, an autoregressive policy can be trained via RL. 
Hence, most recent work on TSP has defaulted to training autoregressive RL models to minimize the tour length \citep{bello2016neural,khalil2017learning,kool2018attention}.
In contrast, \cite{joshi2019efficient} showed that non-autoregressive SL models with sufficient labelled data (generated using Concorde) outperform state-of-the-art RL approaches on fixed graph sizes.
However, their non-autoregressive architecture show poor `zero-shot' generalization performance compared to autoregressive models when evaluated on instances of different sizes than those used for training.

In this paper, we perform controlled experiments on the learning paradigm for autoregressive TSP models\footnote{
We select the autoregressive RL model \citep{kool2018attention} as it naturally fits the sequential nature of TSP and can easily be extended to the SL setting . In contrast, it is not trivial to extend the non-autoregressive SL \citep{joshi2019efficient} model to the RL setting.
}
with an emphasis on \textit{emergent} generalization to variable graph sizes, especially those larger than training graphs.
We find that both SL and RL learn to solve TSP on fixed graph sizes very close to optimal, with SL models achieving state-of-the-art results for TSP20-TSP100. 
Interestingly, RL emerges as the superior learning paradigm for zero-shot generalization to variable and large-scale TSP instances up to TSP500.

Our findings suggest that learning guided by sparse reward functions trains policies which identify more general motifs and patterns for TSP.
In contrast, policies trained to imitate optimal solutions overfit to specific graph sizes. 
We contribute to existing literature on neural combinatorial optimization by empirically exploring 
the impact of learning paradigms for TSP, 
and believe that RL will be a key component towards building scale-invariant solvers for new combinatorial problems beyond TSP.

\section{Experiments}

Intuitively, learning from variable graph sizes is a straightforward way of building more robust and scale-invariant solvers.
In our experiments, we chose to focus on learning from graphs of fixed sizes because we want to study the impact of the learning paradigm on \textit{emergent} generalization in the extreme case; 
where generalization is measured as performance on smaller or larger instances of the same combinatorial problem that the model was trained on.

\paragraph{Model Setup}
We follow the experimental setup of \cite{kool2018attention}\footnote{We modify their codebase: \url{https://github.com/wouterkool/attention-learn-to-route}} to train autoregressive Graph Attention-based models for TSP20, TSP50 and TSP100, and evaluate on instances from TSP20 up till TSP500.
We train two otherwise equivalent variants of the model:
an RL model trained with REINFORCE \citep{williams1992simple} and a greedy rollout baseline; and an SL model trained to minimize a Cross Entropy Loss between the model's predictions and optimal targets at each step, similar to supervised Pointer Networks \citep{vinyals2015pointer}.
We use the model architecture and optimizer specified by \cite{kool2018attention} for both approaches. 
Optimal TSP datasets from \cite{joshi2019efficient} are used to train the SL models whereas training data is generated on the fly for RL. 
See Appendix \ref{app:training} for detailed descriptions of training setups.

\paragraph{Evaluation}
We measure performance on held-out test sets of $10,000$ instances of TSP20, TSP50 and TSP100 from \cite{joshi2019efficient}, 
as well as $1,000$ test instances of TSP150, TSP200, TSP300, TSP400 and TSP500 generated using Concorde.
We use the average predicted tour length and the average optimality gap (percentage ratio of the predicted tour length relative to the optimal solution) over the test sets as performance metrics.
We evaluate both models in three search settings:
greedy search, sampling from the learnt policy ($1,280$ solutions), and beam search (with beam width $1,280$).

\section{Results}

\paragraph{Performance on training graph sizes}
Table \ref{table:main} presents the performance of SL and RL models for various TSP sizes. 
In general, we found that both SL and RL models learn solvers close to optimal for TSP20, TSP50 and TSP100 when trained on the problem size.
In the greedy setting, RL models clearly outperform SL models.
As we sample or perform beam search, SL models obtains state-of-the-art results for all graph sizes, 
showing significant improvement over RL models as well as non-autoregressive SL models from \cite{joshi2019efficient}; \textit{e.g.} TSP100 optimality gap of $0.38\%$ for TSP100 SL model using beam search vs. $2.85\%$ for TSP100 RL model vs. $1.39\%$ reported in \cite{joshi2019efficient}. 

\begin{figure}[t!]
    \centering
    \subfloat[Greedy Search]{
    \includegraphics[width=0.33\textwidth]{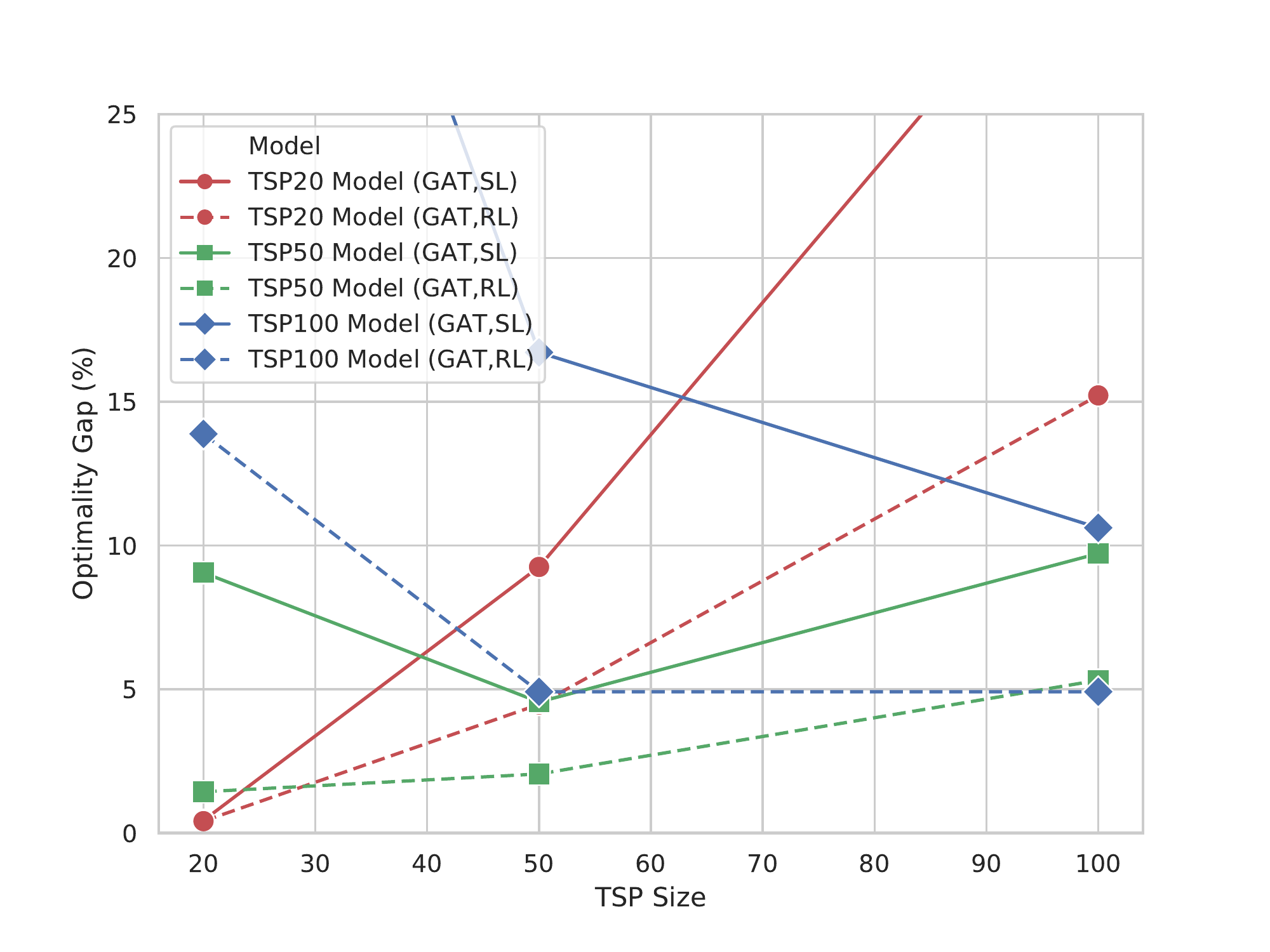}
    \label{fig:main-greedy}
    }
    \subfloat[Sampling]{
    \includegraphics[width=0.33\textwidth]{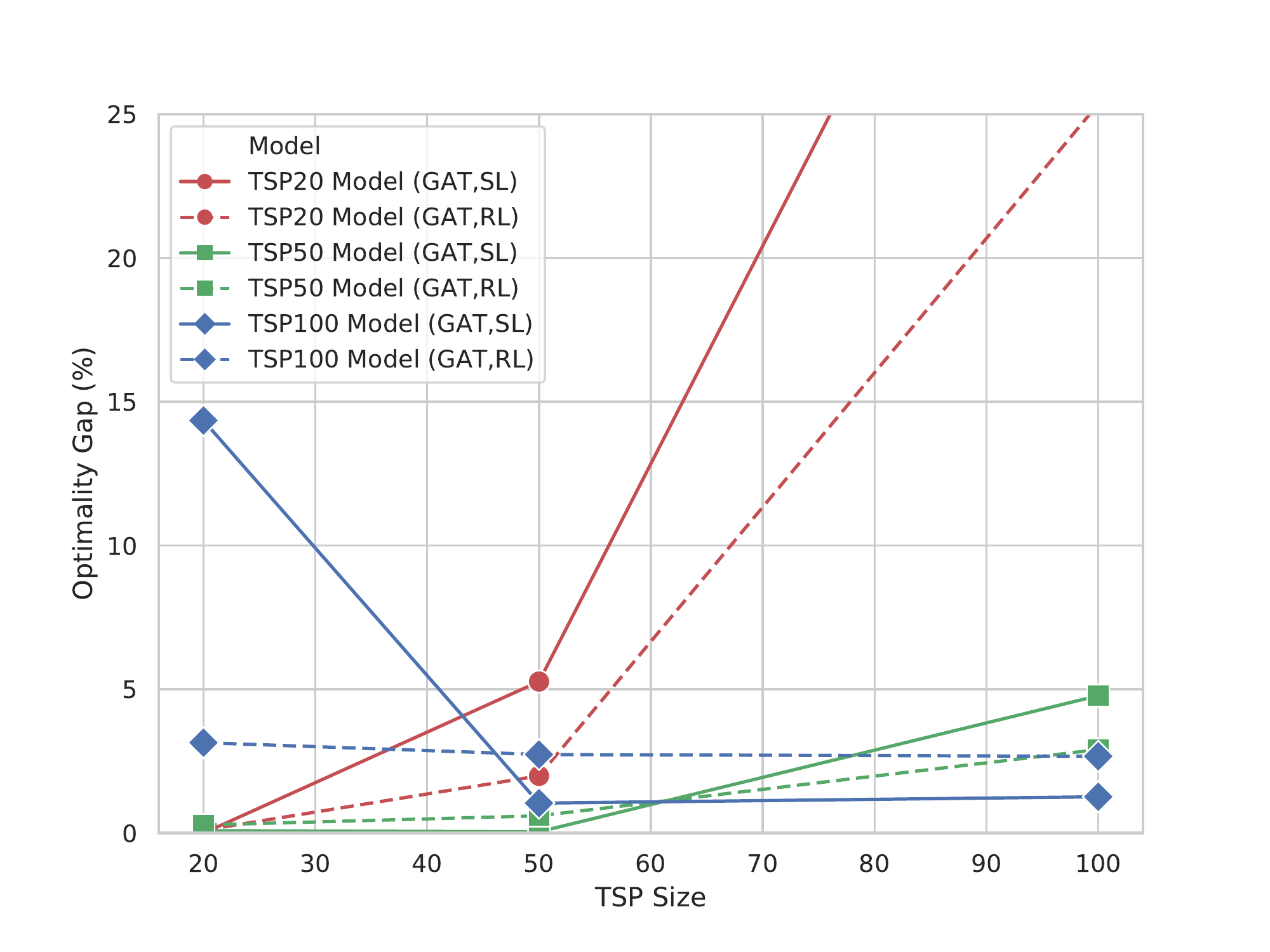}
    \label{fig:main-sample}
    }
    \subfloat[Beam Search]{
    \includegraphics[width=0.33\textwidth]{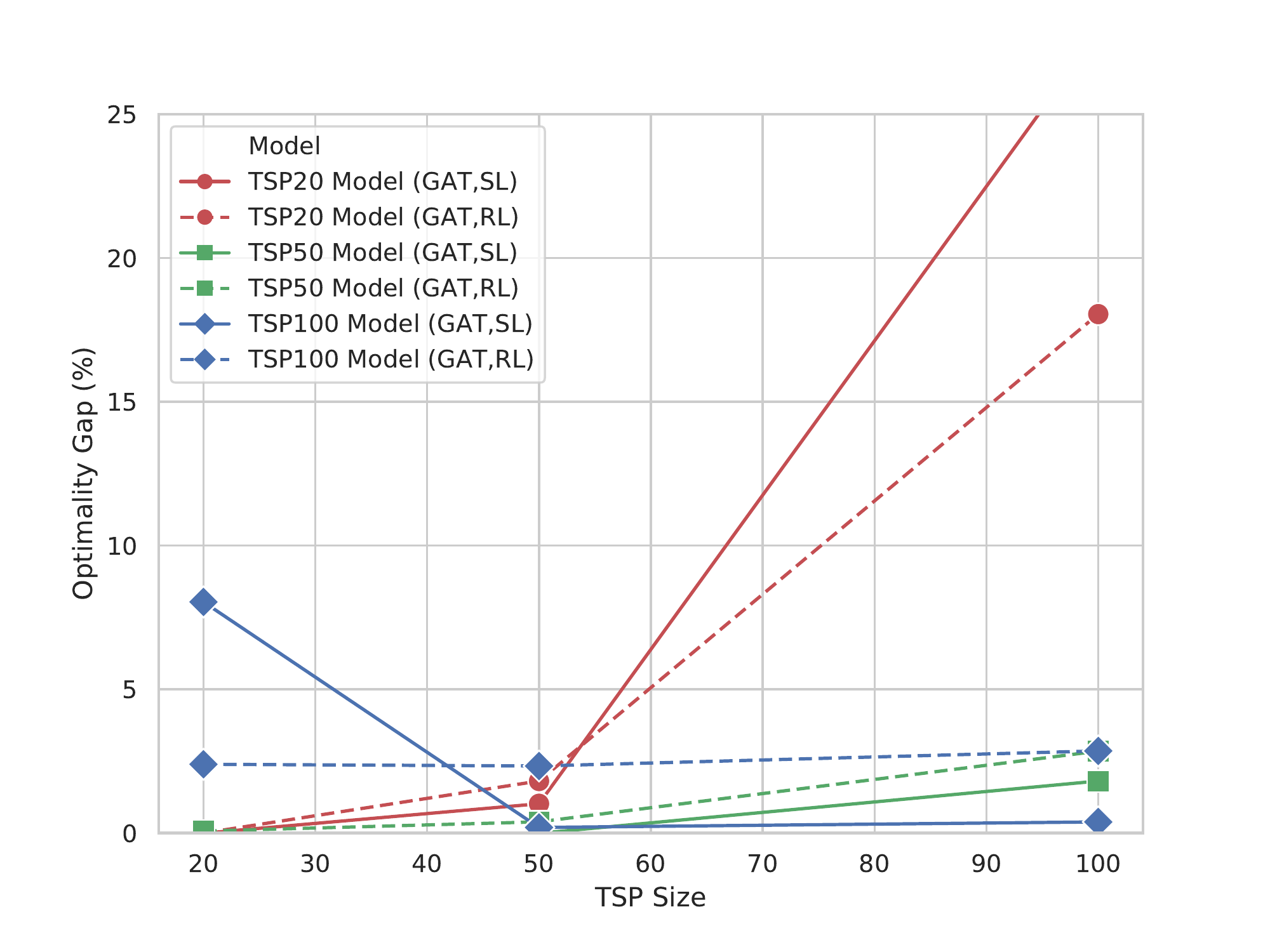}
    \label{fig:main-bs}
    }
    \caption{Zero-shot generalization trends on TSP20-TSP100 for various search settings.}
    \label{fig:main}
\end{figure}

\paragraph{Generalization to variable graph sizes}
RL clearly results in better zero-shot generalization to problem sizes smaller or larger than training graphs.
The different generalization trends for SL and RL models can be visualized in Figure \ref{fig:main} and are highlighted below:
\begin{itemize}
    \item Both TSP20 models do not generalize well to TSP100, but RL training leads to better performance;
    \textit{e.g.} in the greedy setting, TSP100 optimality gap of $32.25\%$ for TSP20 SL model vs. $15.21\%$ for TSP20 RL model.
    \item The TSP50 RL model generalizes to TSP20 and TSP100 better than the TSP50 SL model, except when performing beam search;
    \textit{e.g.} in the sampling setting, TSP100 optimality gap of $4.77\%$ for TSP50 SL model vs. $2.90\%$ for TSP50 RL model.
    \item In all search settings, the TSP100 SL model shows extremely poor generalization to TSP20 compared to the TSP100 RL model;
    \textit{e.g.} in the beam search setting, TSP20 optimality gap of $8.03\%$ for TSP100 SL model vs. $2.39\%$ for TSP100 RL model.
\end{itemize}

TSP solution visualizations for various models are available in Appendix \ref{app:viz}.

\begin{table}[t!]
\centering
\caption{
Test set performance of SL and RL models trained on various TSP sizes.
In the \emph{Type} column, \textbf{SL}: Supervised Learning, \textbf{RL}: Reinforcement Learning. In the \emph{Decoder} column, \textbf{G}: Greedy search, \textbf{S}: Sampling,  \textbf{BS}: Beam search
}
\label{table:main}
\resizebox{\textwidth}{!}{%
\begin{tabular}{lcc|ccc|ccc|ccc}
\toprule
\multirow{2}{*}{Model} & \multirow{2}{*}{Type} & \multirow{2}{*}{Decoder} & \multicolumn{3}{c|}{TSP20} & \multicolumn{3}{c|}{TSP50} & \multicolumn{3}{c}{TSP100} \\
 & & & Tour Len. & Opt. Gap. & Time & Tour Len. & Opt. Gap. & Time & Tour Len. & Opt. Gap. & Time \\
\midrule
Concorde & & &  $3.831$ & $0.000 \%$ & (1m) & $5.692$ & $0.00 \%$ & (2m) & $7.764$ & $0.00 \%$ & (3m) \\
\midrule 
\multicolumn{12}{c}{Greedy search} \\
\midrule
TSP20 Model & SL & G & $3.847$ & $0.437 \%$ & (1s) & $6.219$ & $9.254 \%$ & (1s) & $10.269$ & $32.255 \%$ & (5s) \\
TSP50 Model & SL & G & $4.177$ & $9.052 \%$ & (1s) & $5.951$ & $4.557 \%$ & (1s) & $8.519$ & $9.721 \%$ & (5s) \\
TSP100 Model & SL & G & $5.696$ & $48.712 \%$ & (1s) & $6.643$ & $16.717 \%$ & (1s) & $8.589$ & $10.616 \%$ & (5s) \\
\midrule
TSP20 Model & RL & G & $3.846$ & $0.412 \%$ & (1s) & $5.946$ & $4.473 \%$ & (1s) & $8.946$ & $15.221 \%$ & (5s) \\
TSP50 Model & RL & G & $3.885$ & $1.434 \%$ & (1s) & $5.809$ & $2.053 \%$ & (1s) & $8.177$ & $5.312 \%$ & (5s) \\
TSP100 Model & RL & G & $4.362$ & $13.881 \%$ & (1s) & $5.971$ & $4.911 \%$ & (1s) & $8.146$ & $4.912 \%$ & (5s) \\
\midrule
\multicolumn{12}{c}{Sampling, 1280 solutions} \\
\midrule
TSP20 Model & SL & S & $3.831$ & $0.000 \%$ & (6m) & $5.992$ & $5.268 \%$ & (26m) & $11.115$ & $43.151 \%$ & (1.5h) \\
TSP50 Model & SL & S & $3.834$ & $0.082 \%$ & (6m) & $5.694$ & $0.041 \%$ & (26m) & $8.135$ & $4.776 \%$ & (1.5h) \\
TSP100 Model & SL & S & $4.380$ & $14.348 \%$ & (6m) & $5.751$ & $1.038 \%$ & (26m) & $7.862$ & $1.259 \%$ & (1.5h) \\
\midrule
TSP20 Model & RL & S & $3.834$ & $0.099 \%$ & (6m) & $5.805$ & $1.987 \%$ & (26m) & $9.733$ & $25.357 \%$ & (1.5h) \\
TSP50 Model & RL & S & $3.841$ & $0.278 \%$ & (6m) & $5.726$ & $0.598 \%$ & (26m) & $7.990$ & $2.901 \%$ & (1.5h) \\
TSP100 Model & RL & S & $3.951$ & $3.141 \%$ & (6m) & $5.847$ & $2.730 \%$ & (26m) & $7.972$ & $2.688 \%$ & (1.5h) \\
\midrule
\multicolumn{12}{c}{Beam search, 1280 width} \\
\midrule
TSP20 Model & SL & BS & $3.831$ & $0.000 \%$ & (4m) & $5.750$ & $1.017 \%$ & (30m) & $9.928$ & $27.859 \%$ & (2h) \\
TSP50 Model & SL & BS & $3.831$ & $0.000 \%$ & (4m) & $5.692$ & $0.000 \%$ & (30m) & $7.905$ & $1.807 \%$ & (2h) \\
TSP100 Model & SL & BS & $4.138$ & $8.036 \%$ & (4m) & $5.703$ & $0.193 \%$ & (30m) & $7.794$ & $0.385 \%$ & (2h) \\
\midrule
TSP20 Model & RL & BS & $3.831$ & $0.000 \%$ & (4m) & $5.795$ & $1.808 \%$ & (30m) & $9.166$ & $18.048 \%$ & (2h) \\
TSP50 Model & RL & BS & $3.833$ & $0.067 \%$ & (4m) & $5.714$ & $0.388 \%$ & (30m) & $7.986$ & $2.847 \%$ & (2h) \\
TSP100 Model & RL & BS & $3.922$ & $2.390 \%$ & (4m) & $5.824$ & $2.330 \%$ & (30m) & $7.986$ & $2.855 \%$ & (2h) \\
\bottomrule
\end{tabular}%
}
\end{table}

\paragraph{Performance on large-scale instances}
We further evaluate all SL and RL models on TSP instances with up to $500$ nodes  to determine the upper limits of emergent generalization. 
During sampling/beam search, we only sample/search for $250$ solutions instead of $1,280$ as in previous experiments due to time and memory constraints.

\begin{figure}[h!]
    \centering
    \subfloat[Greedy Search]{
    \includegraphics[width=0.33\textwidth]{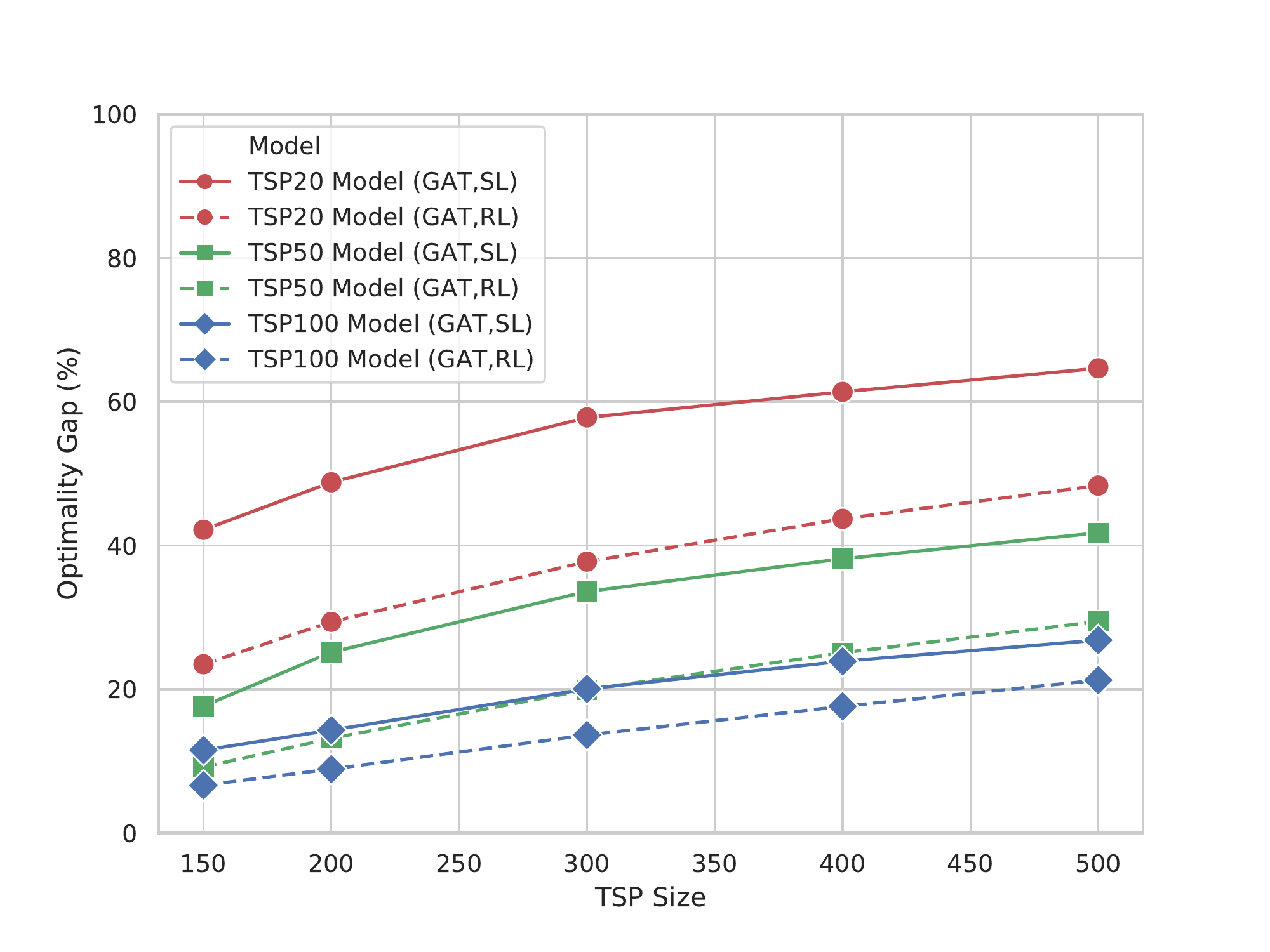}
    \label{fig:large-greedy}
    }
    \subfloat[Sampling]{
    \includegraphics[width=0.33\textwidth]{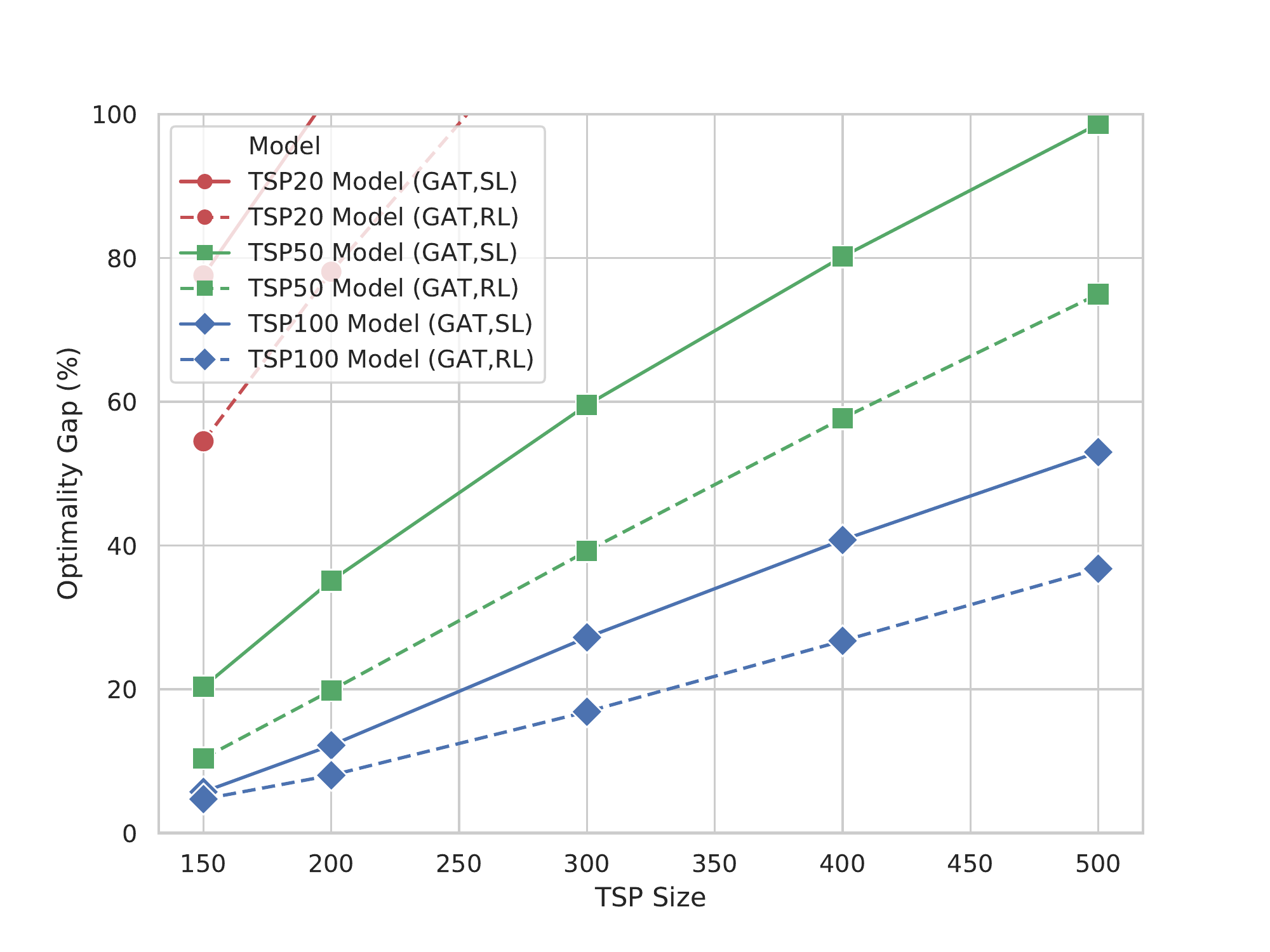}
    \label{fig:large-sample}
    }
    \subfloat[Beam Search]{
    \includegraphics[width=0.33\textwidth]{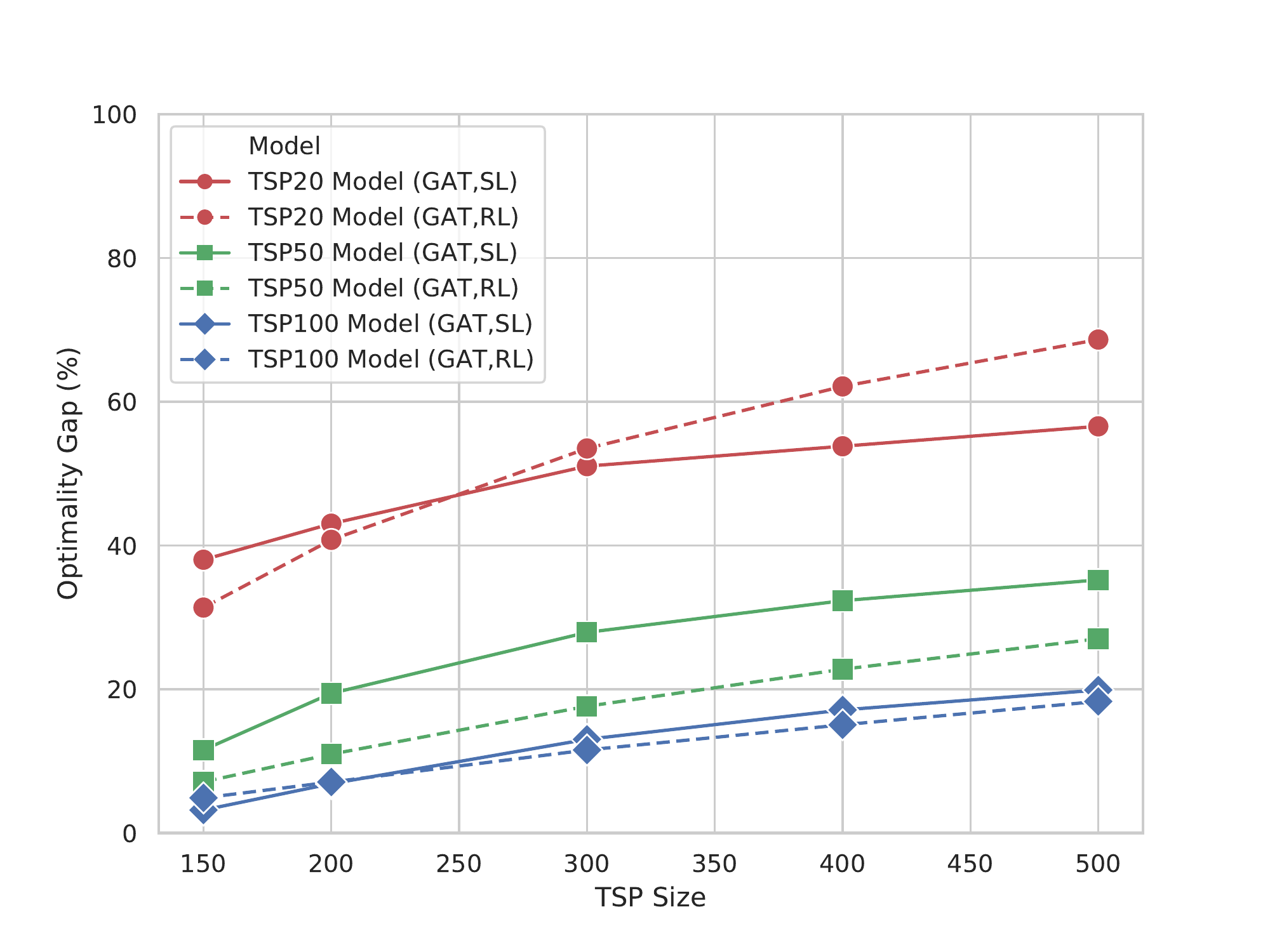}
    \label{fig:large-bs}
    }
    \caption{Zero-shot generalization trends on TSP150-TSP500 for various search settings.}
    \label{fig:large-scale}
\end{figure}

Figure \ref{fig:large-scale} shows that RL training leads to relatively better generalization than SL in all search settings.
Although no model is able to solve large instances close to optimal, generalization performance breaks down smoothly, suggesting that an incremental fine-tuning process using RL might be the key to learning solvers for large-scale TSP. 
Additionally, sampling leads to worse performance than greedy search for larger instances.
This may be due to the probability distributions at each decoding step being close to uniform, \textit{i.e.} the learnt policy is not confident about choosing the next node.

\begin{figure}[h!]
    \centering
    \subfloat[TSP20]{
    \includegraphics[width=0.33\textwidth]{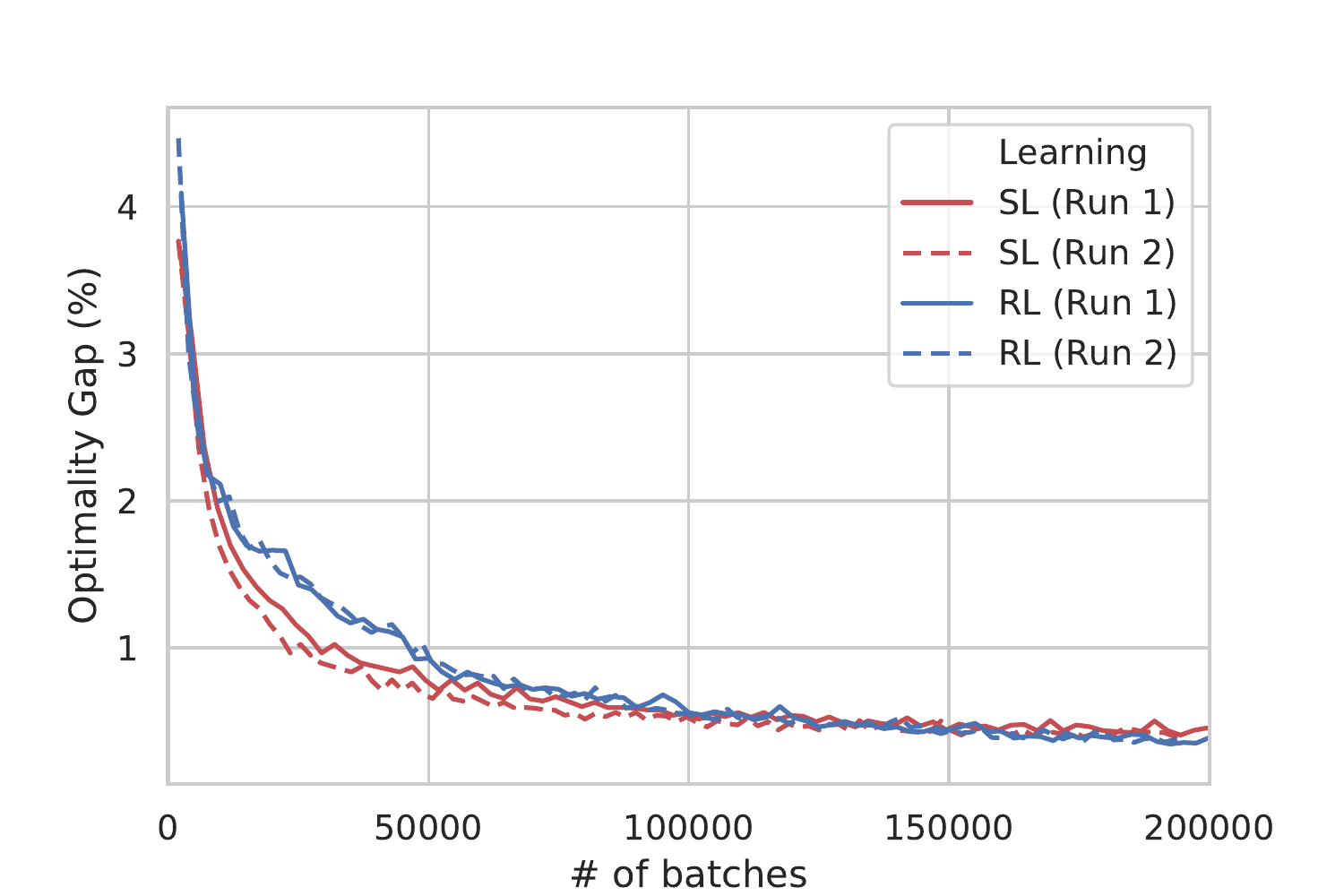}
    \label{fig:train-20}
    }
    \subfloat[TSP50]{
    \includegraphics[width=0.33\textwidth]{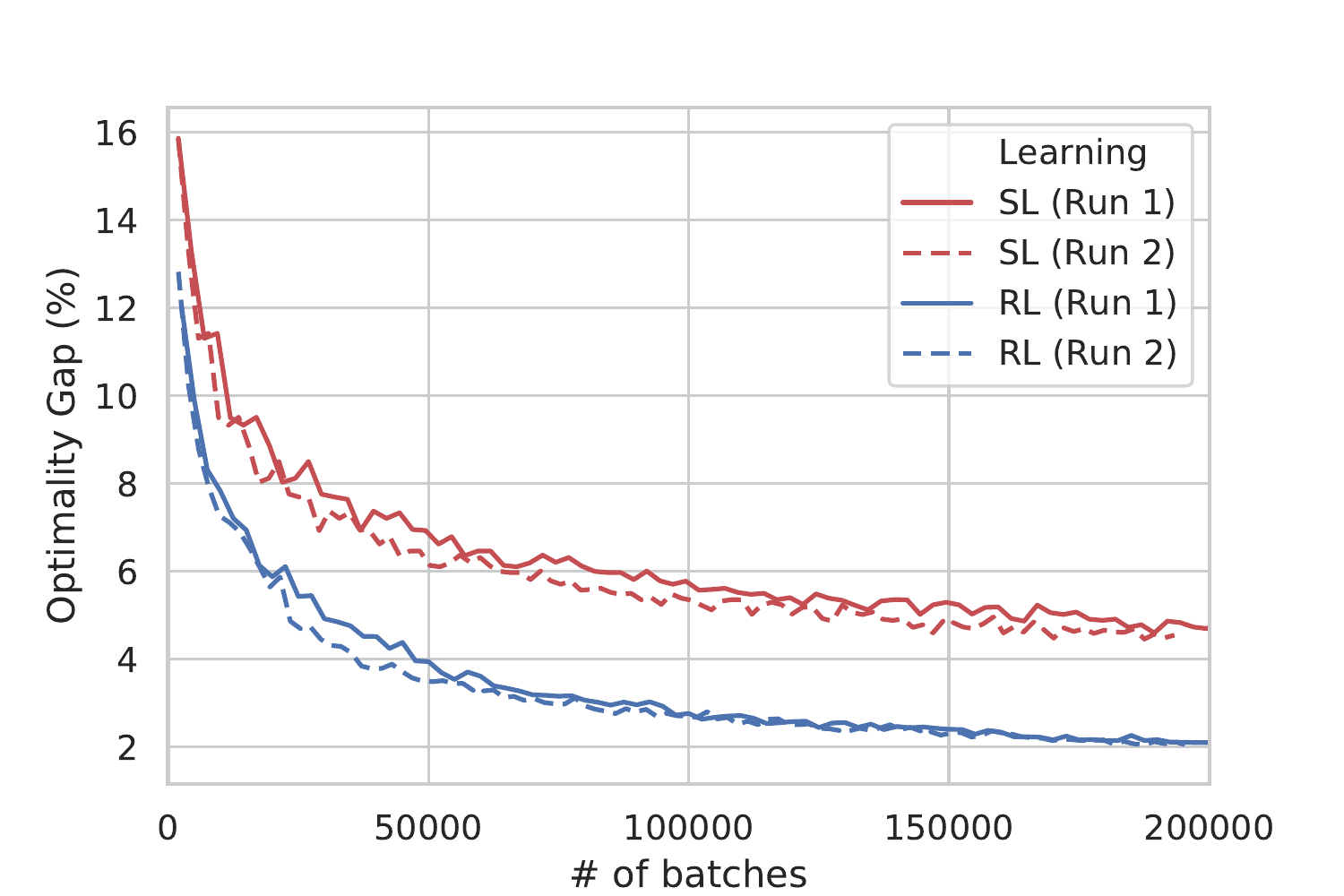}
    \label{fig:train-50}
    }
    \subfloat[TSP100]{
    \includegraphics[width=0.33\textwidth]{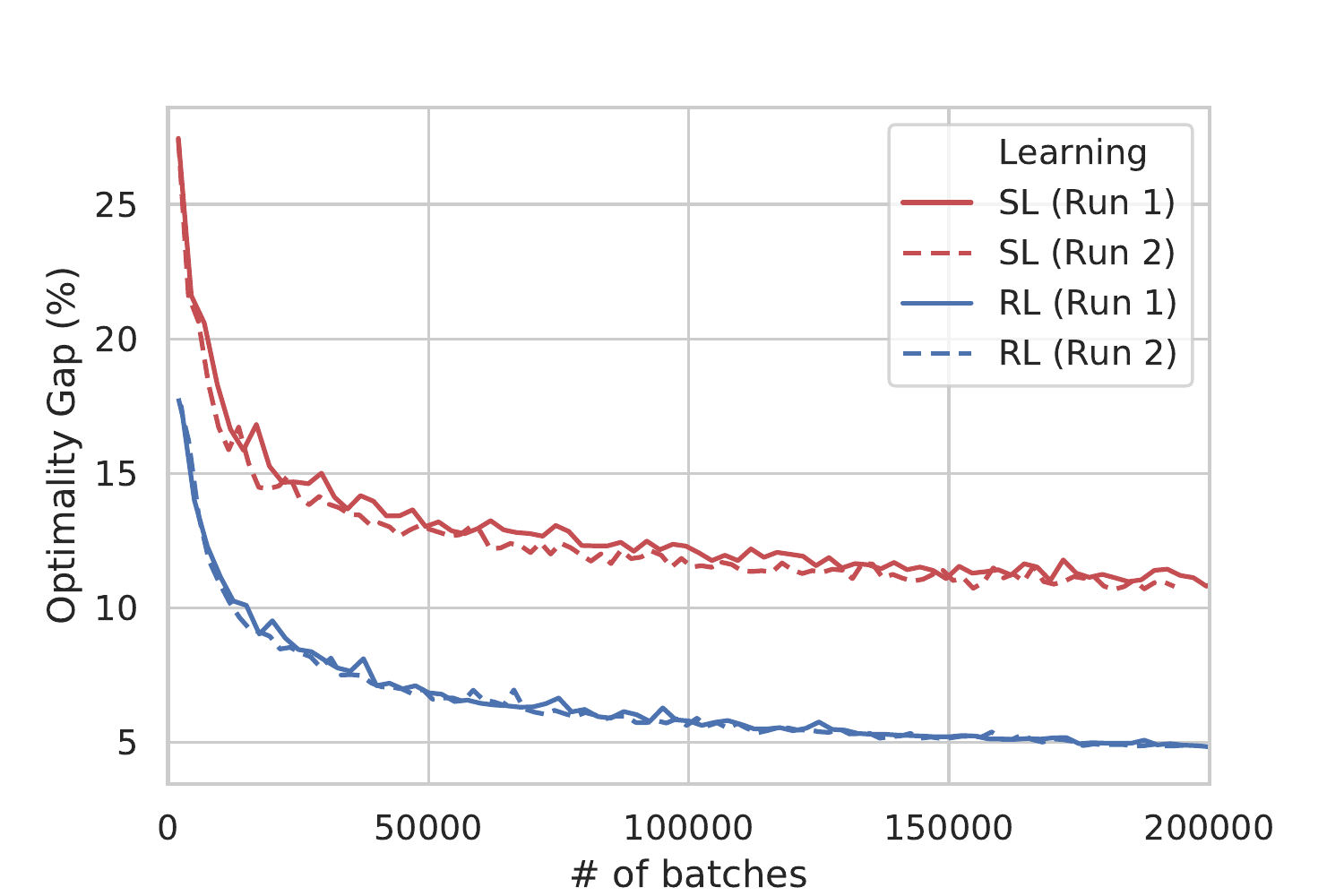}
    \label{fig:train-100}
    }
    \caption{Validation optimality gap (using greedy search) vs. number of training mini-batches for SL and RL models.
    The difference in validation optimality gap is prominent in the greedy setting, but reduces considerably as we sample or use beam search.}
    \label{fig:train}
\end{figure}

\paragraph{Stability and sample efficiency of learning paradigms}
Conventionally, RL is known to be unstable and less sample efficient than SL.
As seen by Figure \ref{fig:train}, we found training to be stable and equally sample efficient in our experiments for both learning paradigms.
SL and RL models require approximately equal number of mini-batches to converge to stable states using a fixed learning rate of $10^{-4}$.
As noted by \cite{kool2018attention}, using larger learning rates with a decay schedule speeds up learning at the cost of stability.

\paragraph{Impact of RL baseline and graph encoder}
In Appendix \ref{app:baseline}, we replace the rollout baseline for REINFORCE with a critic baseline similar to \cite{bello2016neural}. 
We find that RL models follow similar generalization trends to variable sized instances, regardless of the choice of baseline used.

Further, in Appendix \ref{app:gcn-vs-gat}, we replace the \textit{Graph Transformer} encoder with the \textit{Graph Convolutional Network} encoder from \cite{joshi2019efficient}.
We find that choice of graph encoder has negligible impact on performance,
and better generalization in RL models is due to the learning paradigm.

\section{Conclusion}
This paper investigates the choice of learning paradigms for the Travelling Salesman Problem.
Through controlled experiments, we find that both supervised learning and reinforcement learning can train models which perform close to optimal on fixed graph sizes.
Evaluating models on variable sized instances larger than those seen in training reveals a threefold advantage of RL training:
(1) training does not require expensive labelled data;
(2) learning is as stable and sample-efficient as SL;
and most importantly,
(3) RL training leads to better \textit{emergent} generalization on variable sized graphs.
Finding (3) has broader implications on building scale-invariant solvers for practical combinatorial problems beyond TSP.

Future work shall explore more detailed evaluations of learning paradigms for other problems, as well as performing fine-tuning/transfer learning for generalization to large-scale instances.

\section*{Acknowledgement}
XB is supported in part by NRF Fellowship NRFF2017-10.

\bibliographystyle{abbrv}
\bibliography{main}

\newpage
\appendix
\section{Experimental Setup}
\label{app:training}

Table \ref{table:setup} presents an overview of our experimental setup for training autoregressive SL and RL models on fixed graphs.
We use the model architecture specified in \cite{kool2018attention} for both approaches:
The graph encoder is a $3$-layer \textit{Graph Transformer} \citep{velickovic2018graph} with $128$-dimensional embeddings/hidden states and $8$ attention heads per layer. 
The node and graph embeddings produced by the encoder are provided to an autoregressive Attention-based decoder which outputs TSP tours node-by-node. 
Both models are trained using the Adam optimizer \citep{kingma2015adam} with a fixed learning rate of $10^{-4}$ for $100$ epochs with mini-batches of size $512$, where the epoch size is $1,000,000$.

\begin{table}[h!]
\centering
\caption{
Training setup for comparing autoregressive SL and RL models.
}
\label{table:setup}
\begin{tabular}{l|c|c}
\toprule
Parameter & Reinforcement Learning & Supervised Learning \\
\midrule
Training Epochs & $100$ & $100$ \\
Epoch Size & $1,000,000$ & $1,000,000$ \\
Batch Size & $512$ & $512$ \\
Graph Generation & Random & Fixed set of 1M \\
\midrule
Graph Encoder & Graph Transformer & Graph Transformer \\
Encoder Layers & $3$ & $3$ \\
Embedding Dimension & $128$ & $128$ \\
Hidden Dimension & $128$ & $128$ \\
Feed-forward Dimension & $512$ & $512$ \\
Attention Heads & $8$ & $8$ \\
Number of Parameters & $706,608$ & $706,608$ \\
\midrule
Loss Function & REINFORCE, rollout baseline & Cross Entropy Loss \\
Other Tricks & Baseline update after validation & Teacher Forcing \\
Optimizer & Adam & Adam \\
Learning Rate & $10^{-4}$ (fixed) & $10^{-4}$ (fixed) \\
\bottomrule
\end{tabular}%
\end{table}

\section{Replacing Rollout with Critic Baseline}
\label{app:baseline}

\begin{figure}[h!]
    \centering
    \subfloat[Greedy Search]{
    \includegraphics[width=0.33\textwidth]{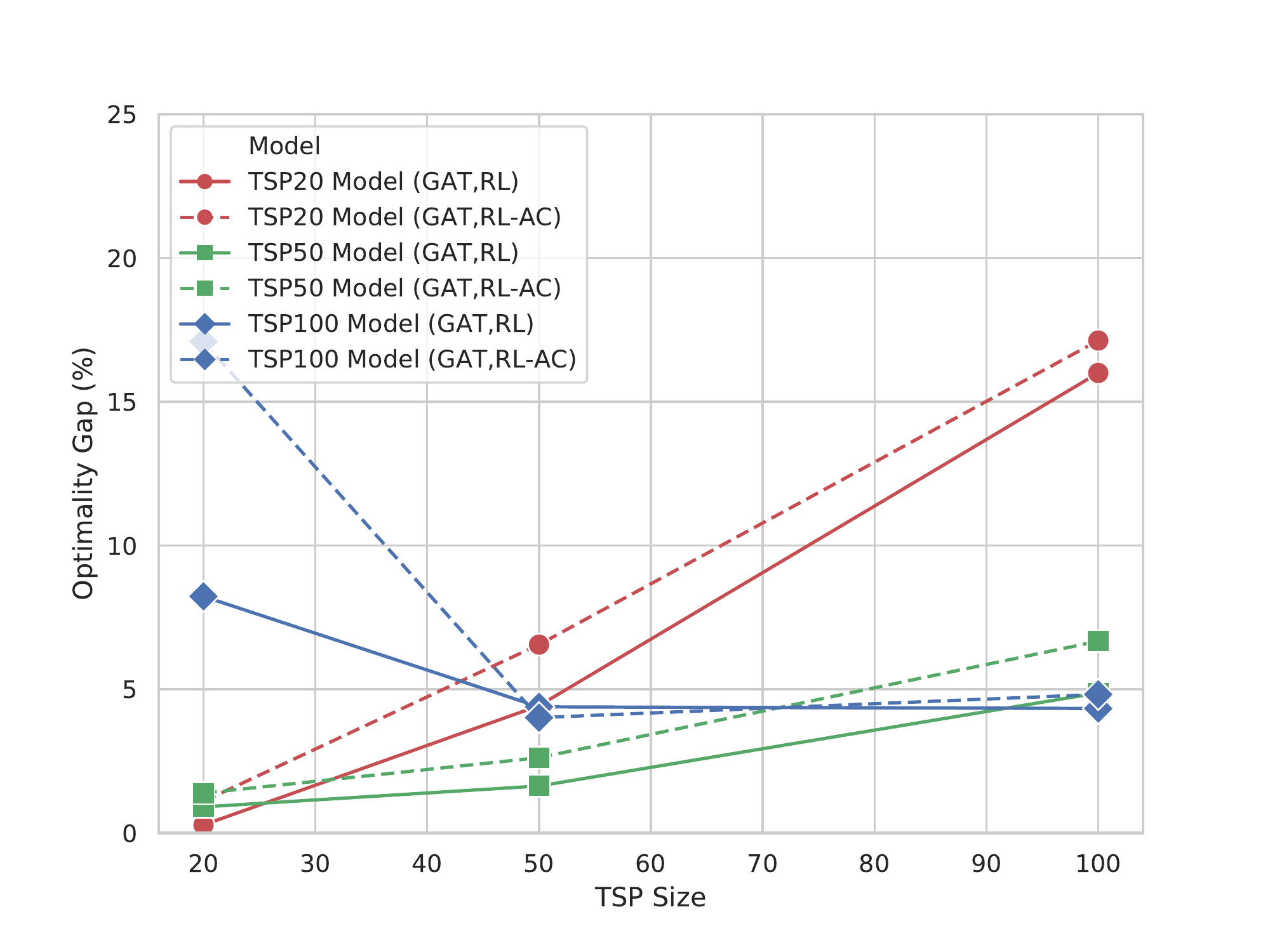}
    \label{fig:rl-greedy}
    }
    \subfloat[Sampling]{
    \includegraphics[width=0.33\textwidth]{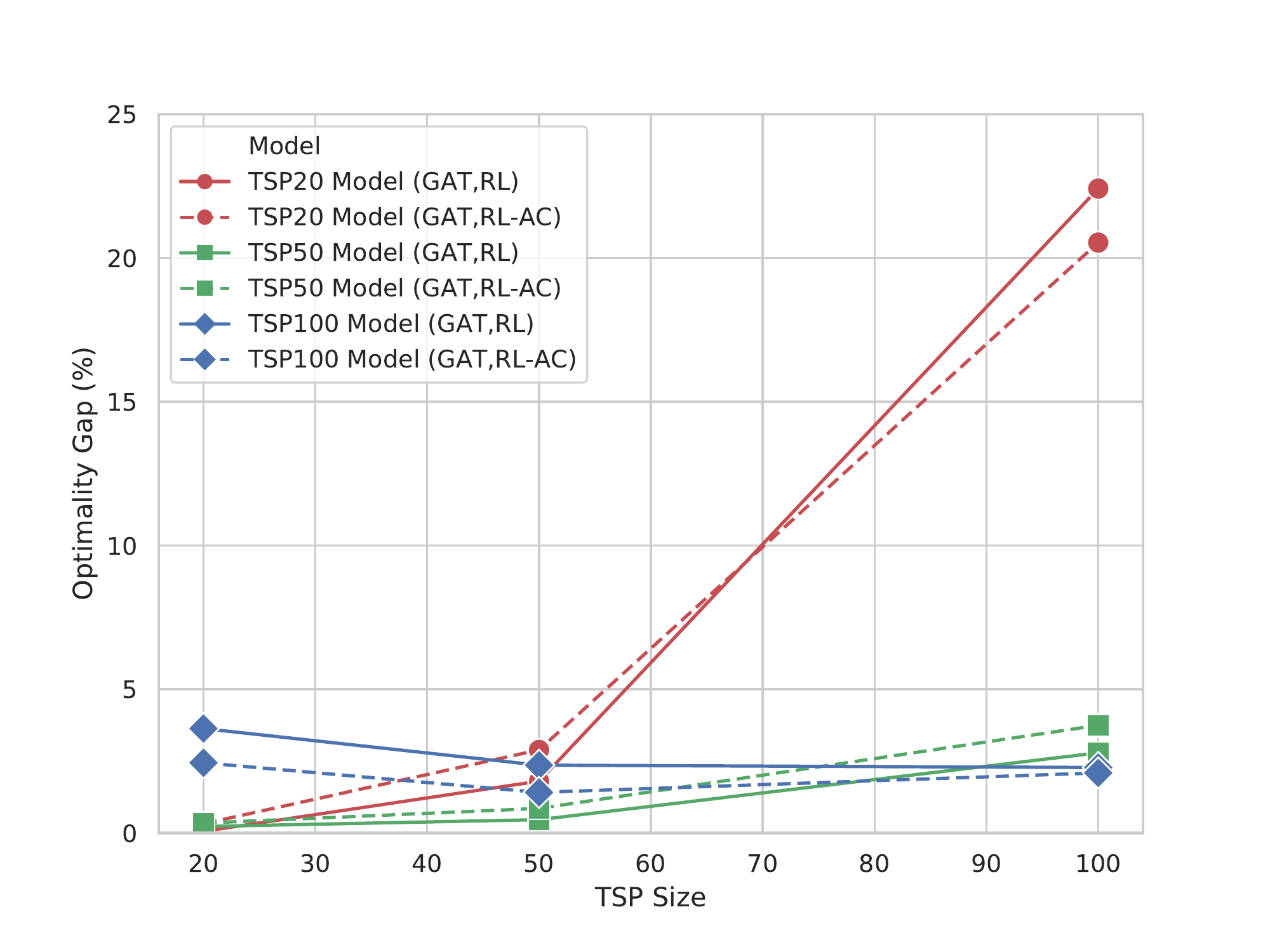}
    \label{fig:rl-sample}
    }
    \subfloat[Beam Search]{
    \includegraphics[width=0.33\textwidth]{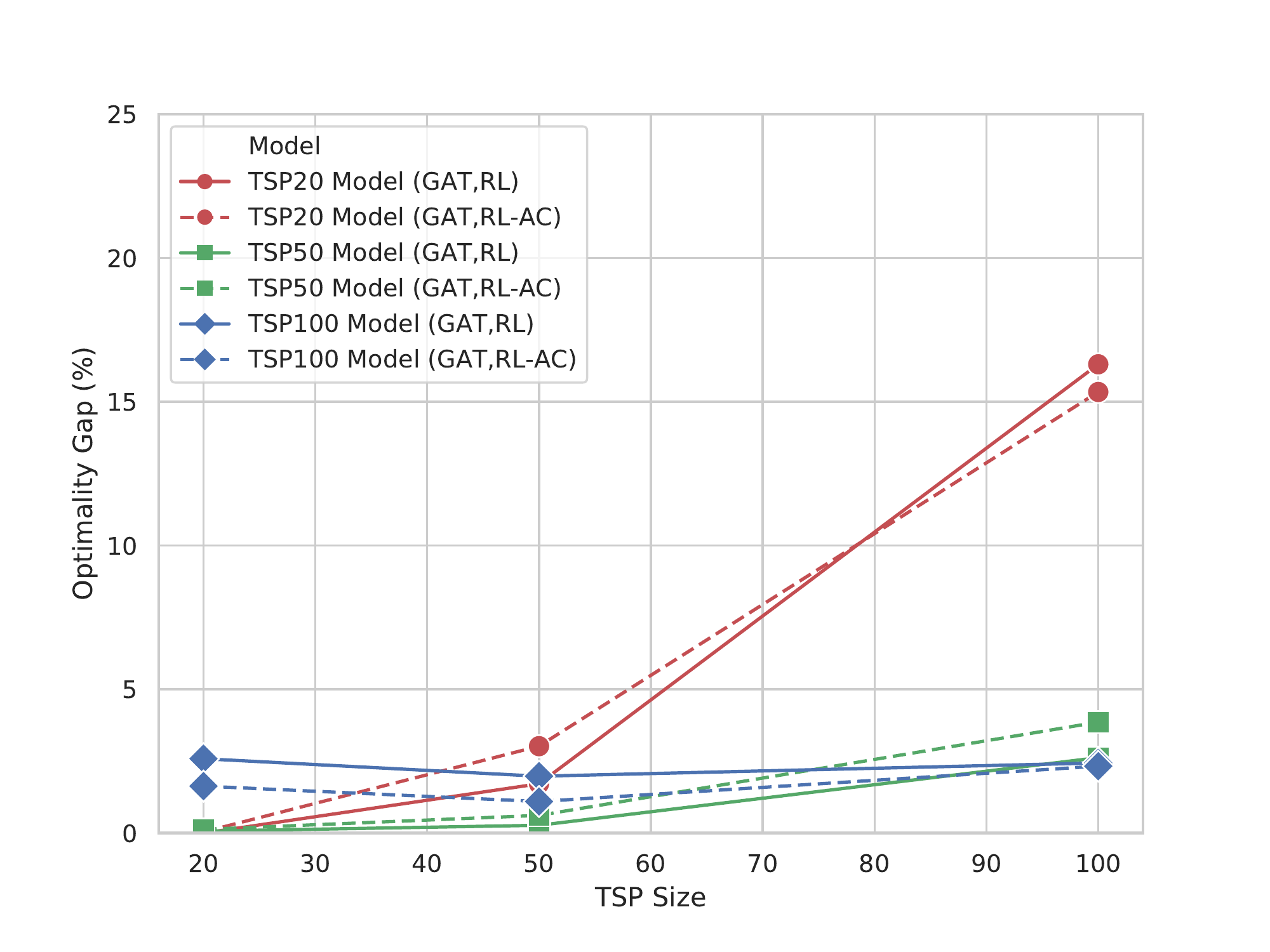}
    \label{fig:rl-bs}
    }
    \caption{Zero-shot generalization trends on TSP20-TSP100 for various search settings.}
    \label{fig:rl-type}
\end{figure}

In Figure \ref{fig:rl-type}, we compare the generalization trends for RL models trained with rollout baselines (as in \cite{kool2018attention}) and critic baselines (as in \cite{bello2016neural}).
The critic network architecture uses the same $3$-layer graph encoder architecture as the main model, after which the node
embeddings are averaged and provided to an MLP with one hidden layer of $128$-units and $\text{ReLU}$ activation, and a single output.
All other experimental settings are kept consistent across both approaches, as in Table \ref{table:setup}. 
We find that RL models follow similar generalization trends to variable sized instances, regardless of the choice of baseline used to train them.

\newpage

\section{Replacing GAT Graph Encoder with GCN}
\label{app:gcn-vs-gat}

To further isolate the impact of learning paradigms on performance and generalization,
we swap the \textit{Graph Transformer} encoder used in \cite{kool2018attention} with the \textit{Graph Convolutional Network} encoder from \cite{joshi2019efficient}, keeping the autoregressive decoder the same.
Our GCN architecture consists of 3 graph convolution layers with $128$-dimensional embeddings/hidden states. 
We use residual connections, batch normalization and edge gates between each layer, as described in \cite{bresson2018experimental}.

\begin{figure}[h!]
    \centering
    \subfloat[TSP20 Model, Greedy Search]{
    \includegraphics[width=0.33\textwidth]{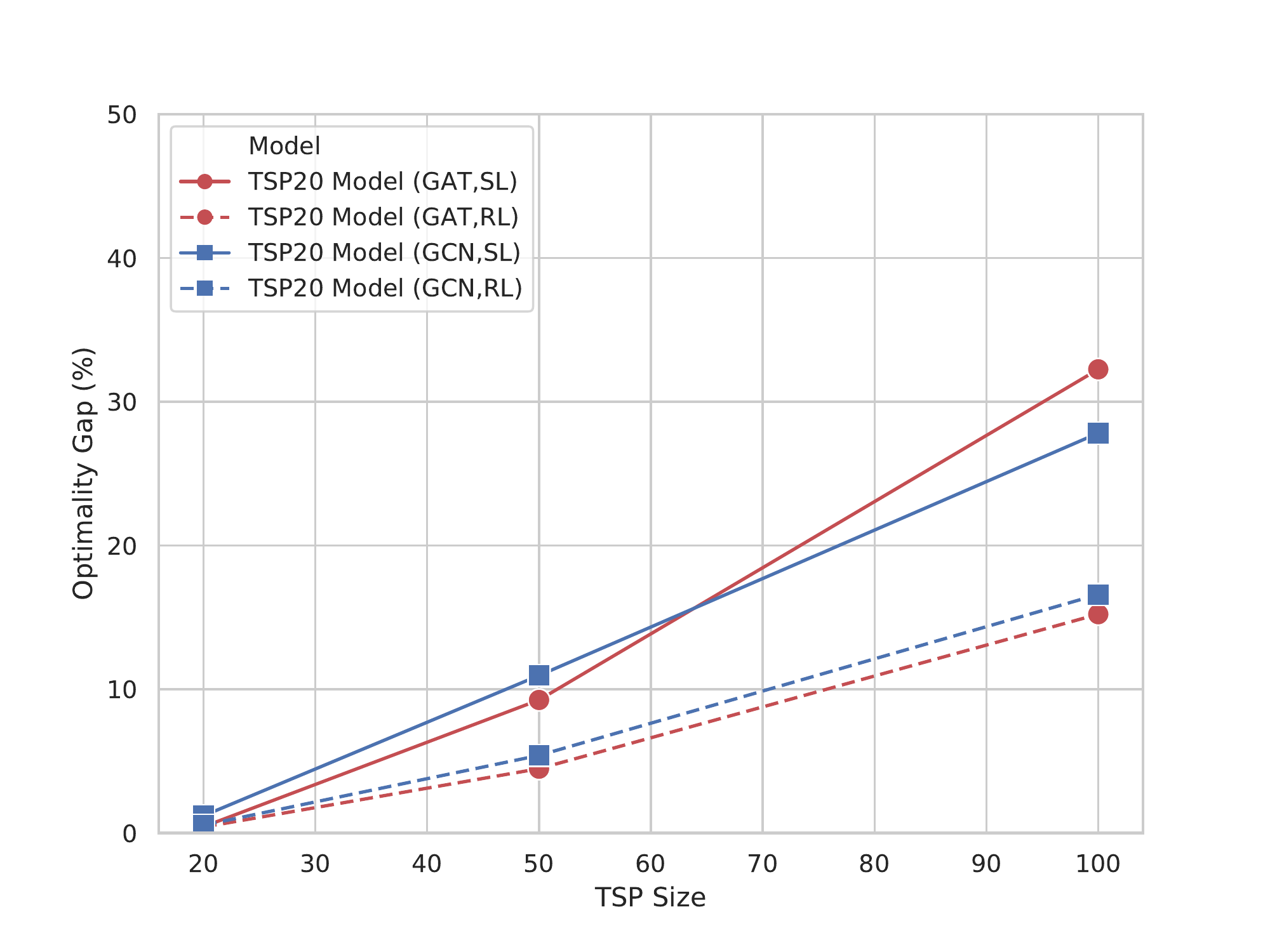}
    \label{fig:gcn20-greedy}
    }
    \subfloat[TSP20 Model, Sampling]{
    \includegraphics[width=0.33\textwidth]{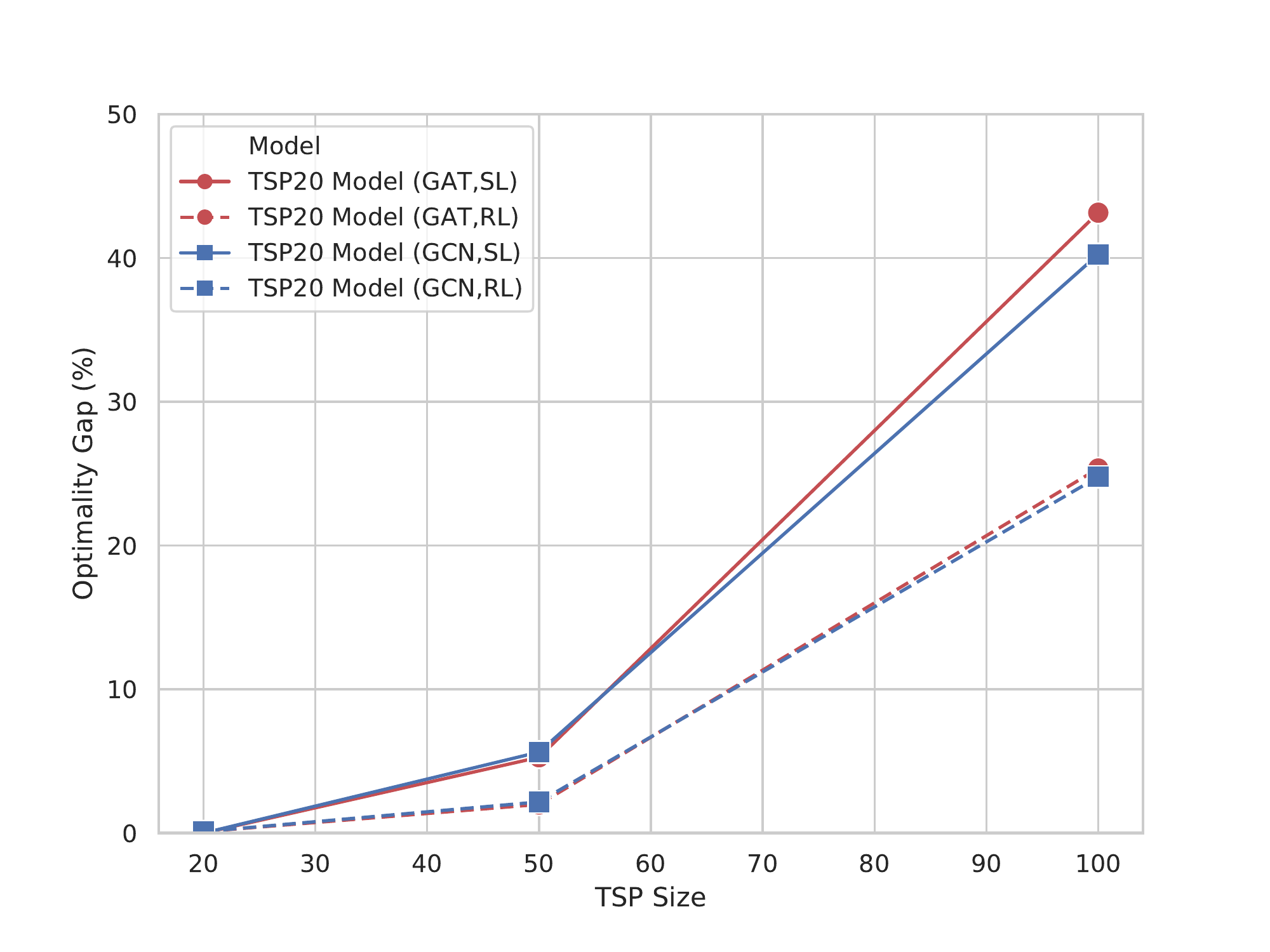}
    \label{fig:gcn20-sample}
    }
    \subfloat[TSP20 Models, Beam Search]{
    \includegraphics[width=0.33\textwidth]{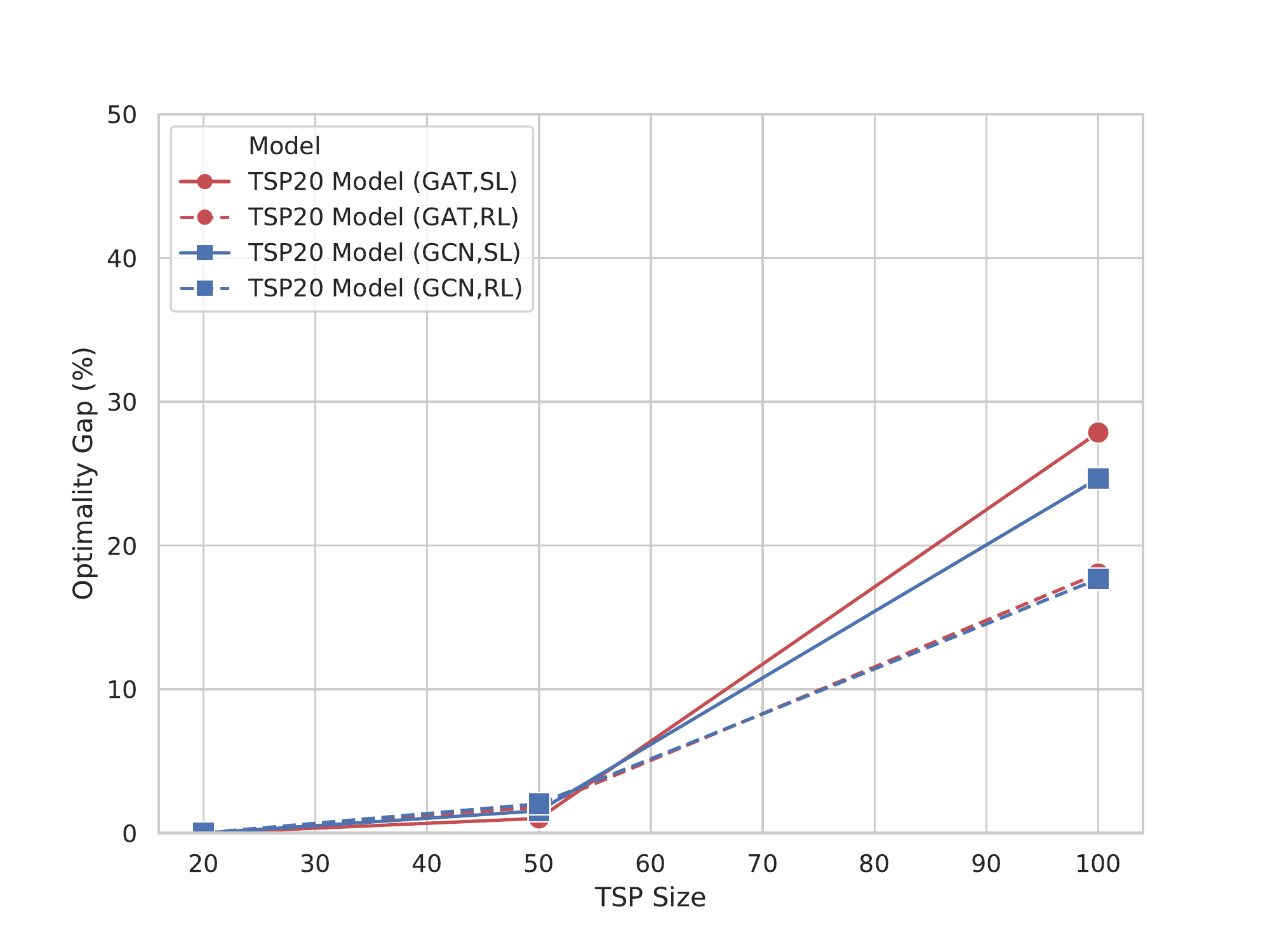}
    \label{fig:gcn20-bs}
    }
    \\
    \subfloat[TSP50 Model, Greedy Search]{
    \includegraphics[width=0.33\textwidth]{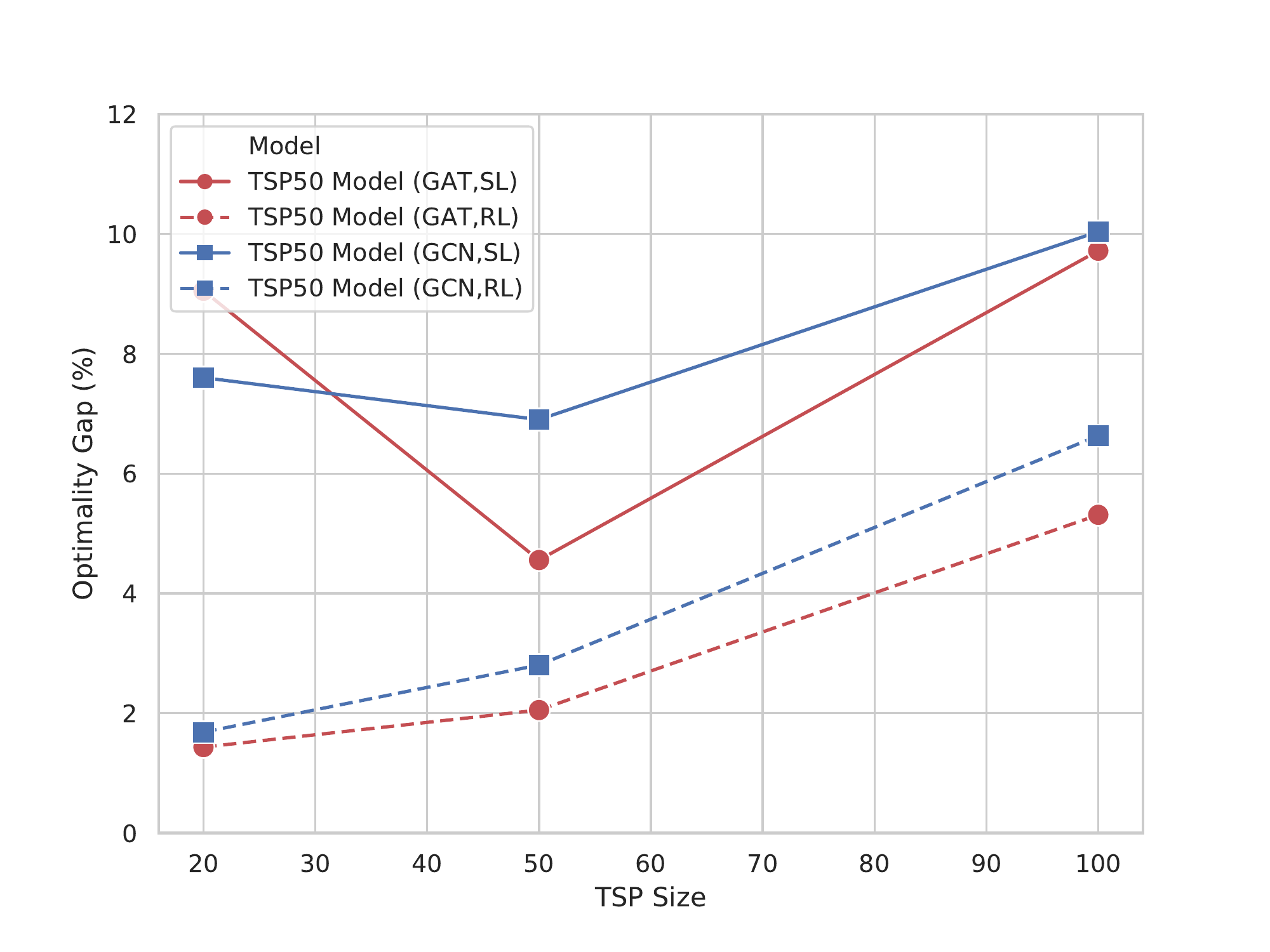}
    \label{fig:gcn50-greedy}
    }
    \subfloat[TSP50 Model, Sampling]{
    \includegraphics[width=0.33\textwidth]{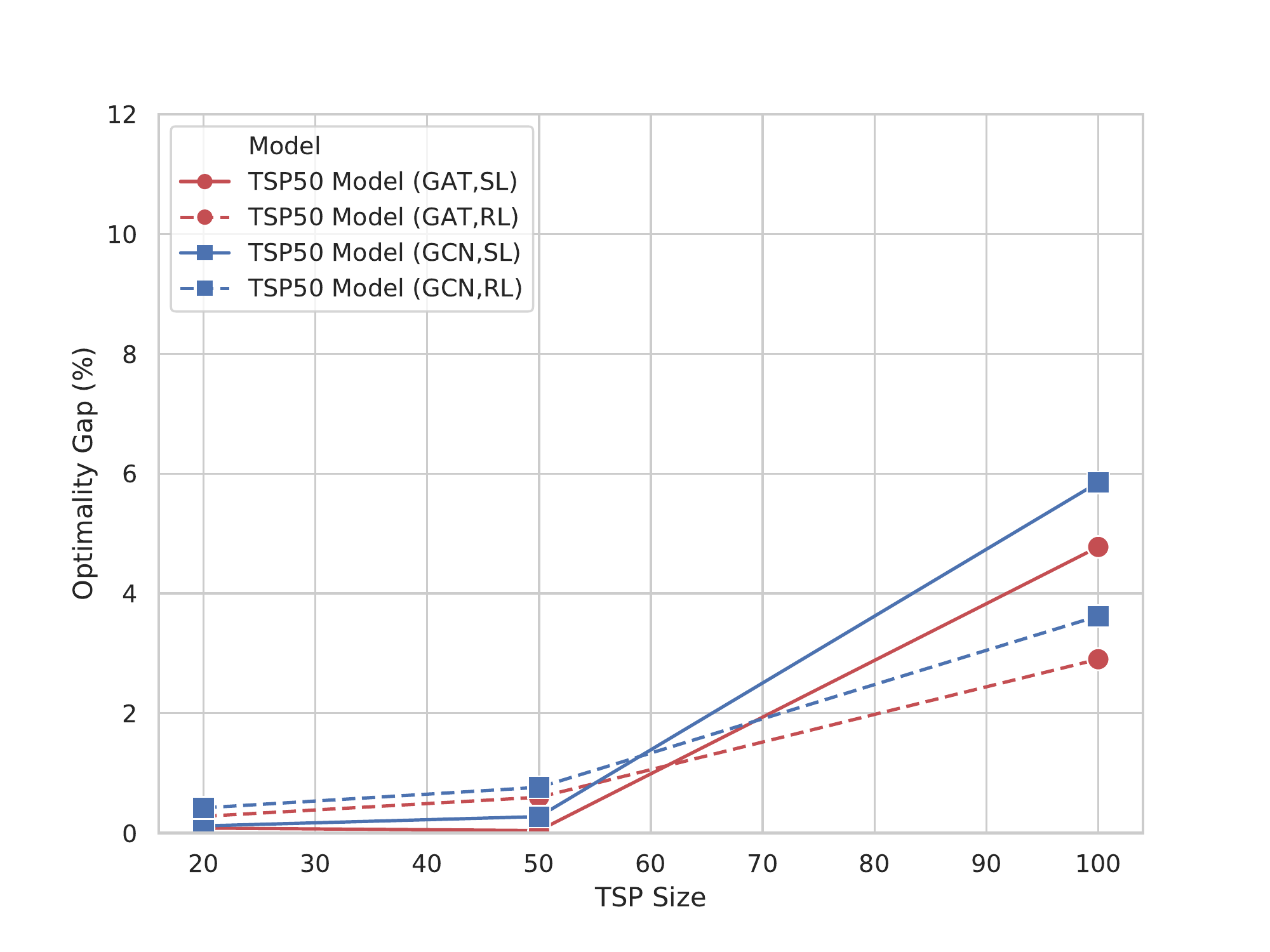}
    \label{fig:gcn50-sample}
    }
    \subfloat[TSP50 Model, Beam Search]{
    \includegraphics[width=0.33\textwidth]{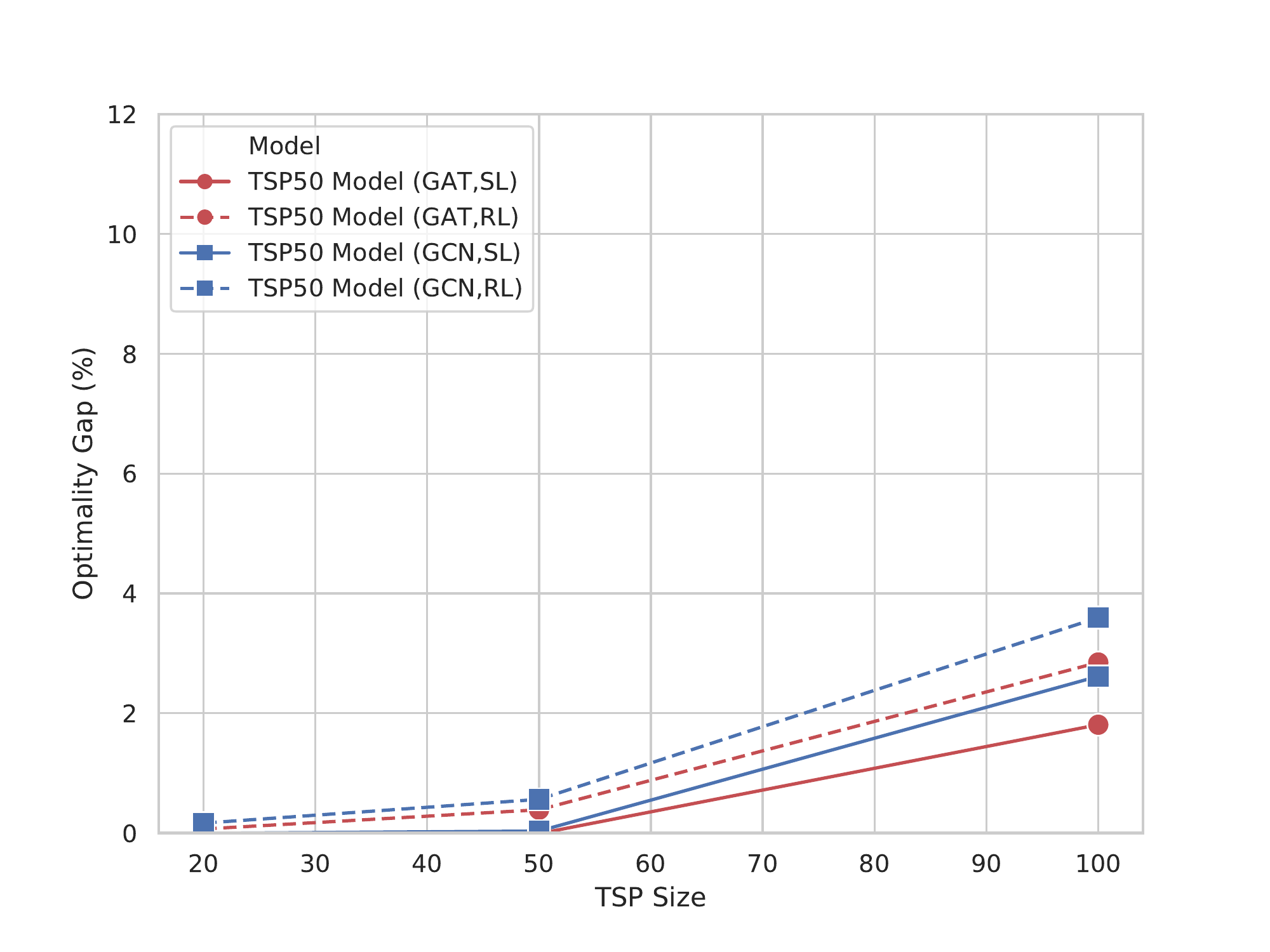}
    \label{fig:gcn50-bs}
    }
    \\
    \subfloat[TSP100 Model, Greedy Search]{
    \includegraphics[width=0.33\textwidth]{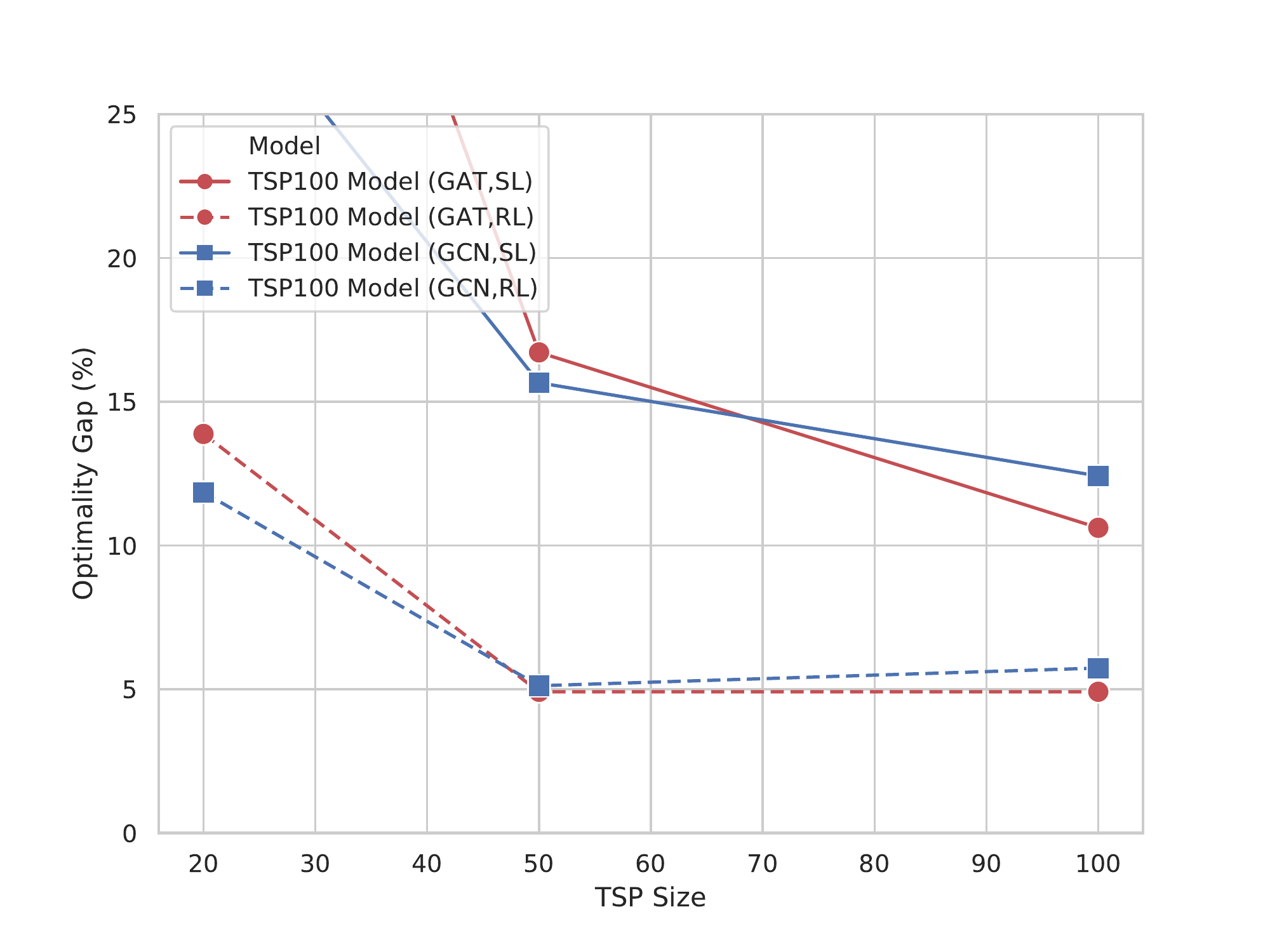}
    \label{fig:gcn100-greedy}
    }
    \subfloat[TSP100 Model, Sampling]{
    \includegraphics[width=0.33\textwidth]{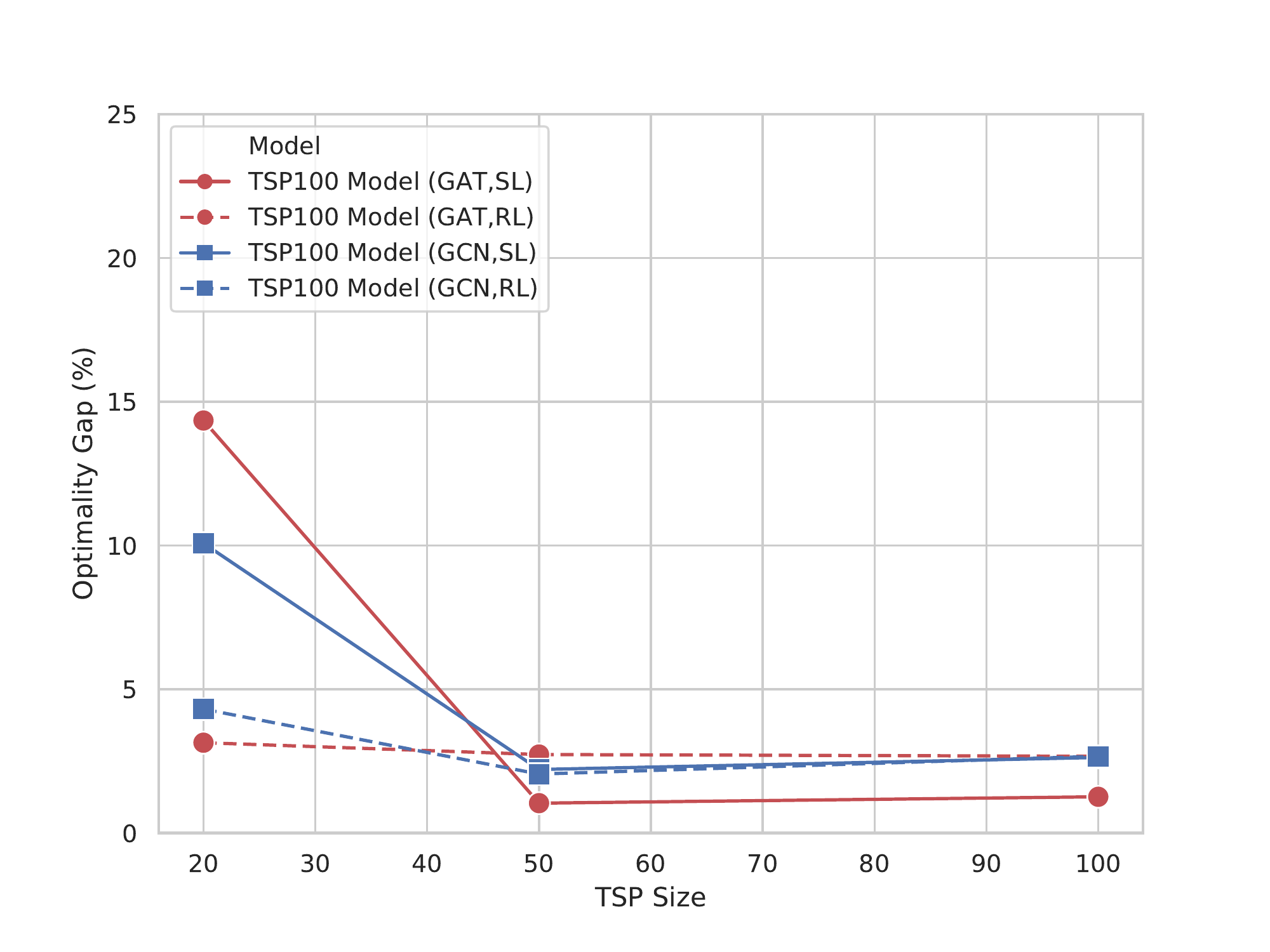}
    \label{fig:gcn100-sample}
    }
    \subfloat[TSP100 Model, Beam Search]{
    \includegraphics[width=0.33\textwidth]{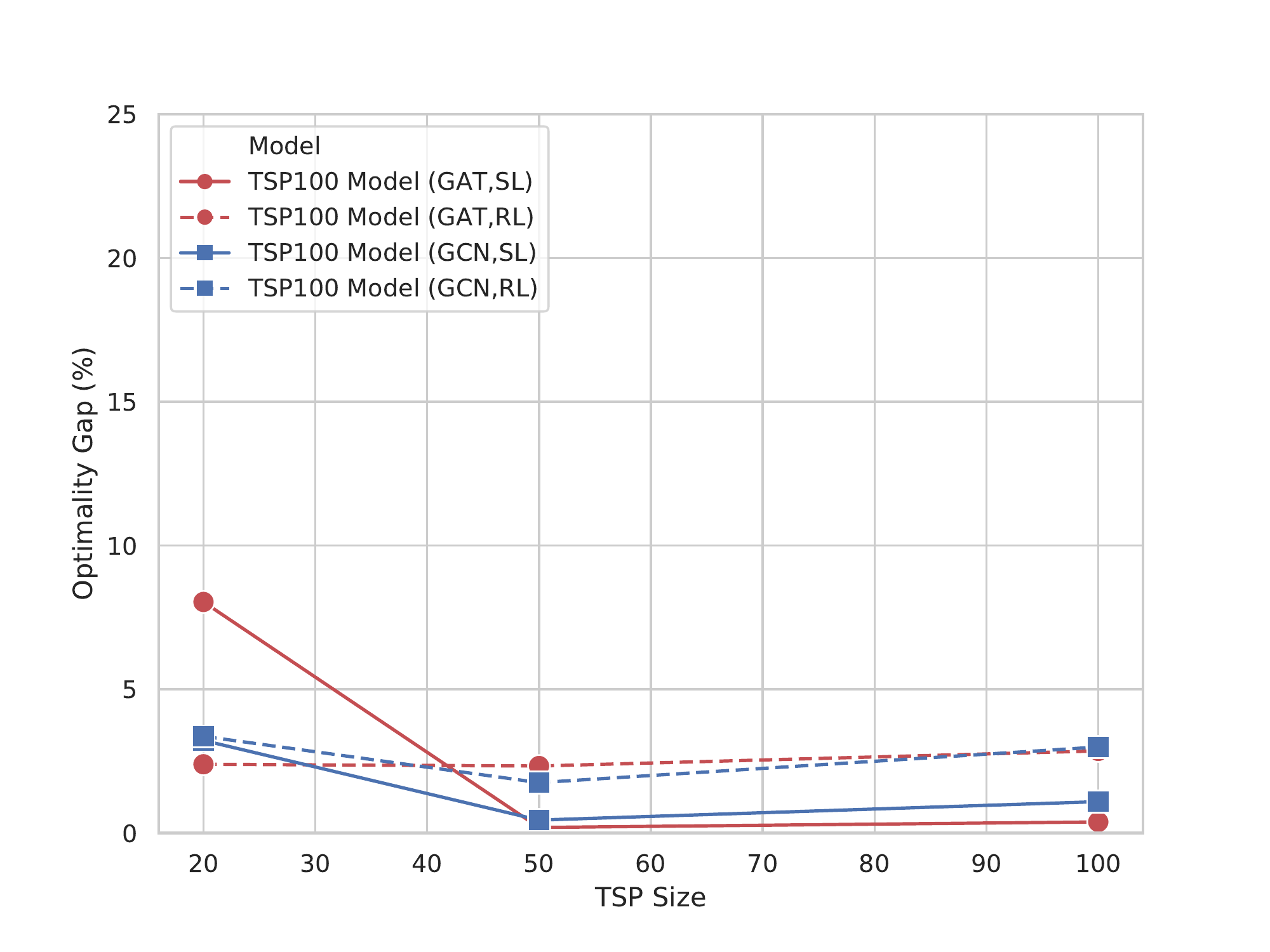}
    \label{fig:gcn100-bs}
    }
    \caption{Zero-shot generalization trends for various search settings for GAT and GCN models.}
    \label{fig:gcn}
\end{figure}

Figure \ref{fig:gcn} plots performance of SL and RL models using GAT and GCN as graph encoders.
We find negligible impact on performance for both learning paradigms
despite GCN models have half the number of parameters as GAT models ($315,264$ for GCN vs. $708,608$ for GAT).\footnote{
We did not experiment with larger GCN encoders as $3$-layer GCNs took longer to train than $3$-layer GATs and had high GPU overhead due to the computation of edge features for fully-connected graphs.
}
Generalization to instances sizes different from training graphs indicates that better emergent generalization in RL models is chiefly due to the learning paradigm, and is independent of the choice of graph encoder.

\newpage

\section{Solution Visualizations}
\label{app:viz}

\begin{figure}[h!]
    \centering
    \subfloat[TSP20 instance, Greedy search]{
    \includegraphics[width=\textwidth]{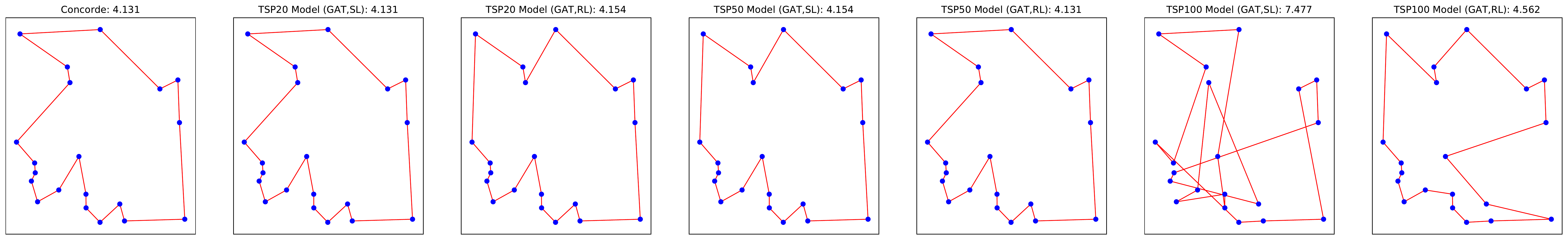}
    \label{fig:viz20-greedy}
    }
    \\
    \subfloat[TSP20 instance, Sampling (1280 solutions)]{
    \includegraphics[width=\textwidth]{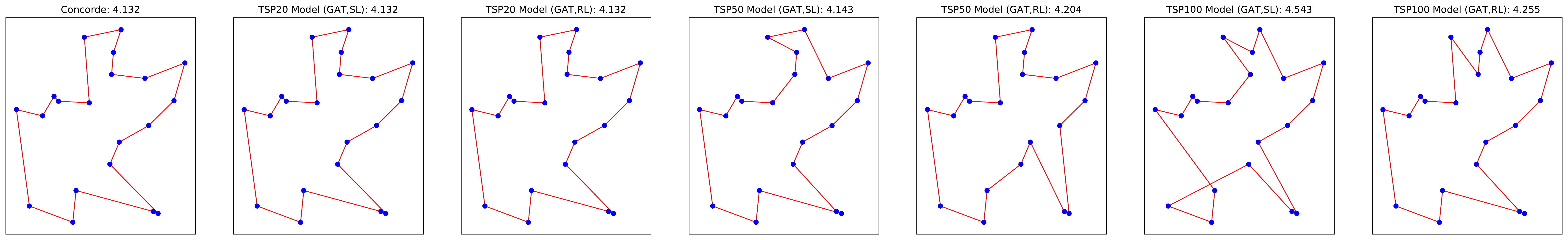}
    \label{fig:viz20-sam}
    }
    \\
    \subfloat[TSP20 instance, Beam search (1280 width)]{
    \includegraphics[width=\textwidth]{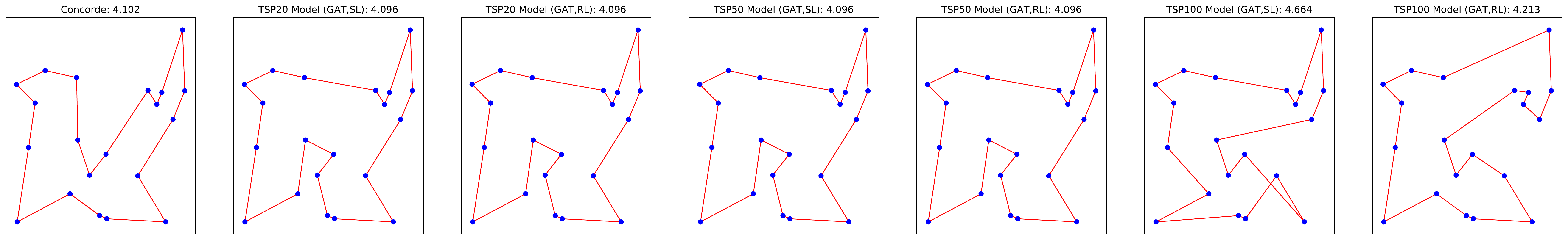}
    \label{fig:viz20-bs}
    }
    \caption{Prediction visualizations from various models for TSP20 test instances.}
    \label{fig:viz20}
\end{figure}

\begin{figure}[h!]
    \centering
    \subfloat[TSP50 instance, Greedy search]{
    \includegraphics[width=\textwidth]{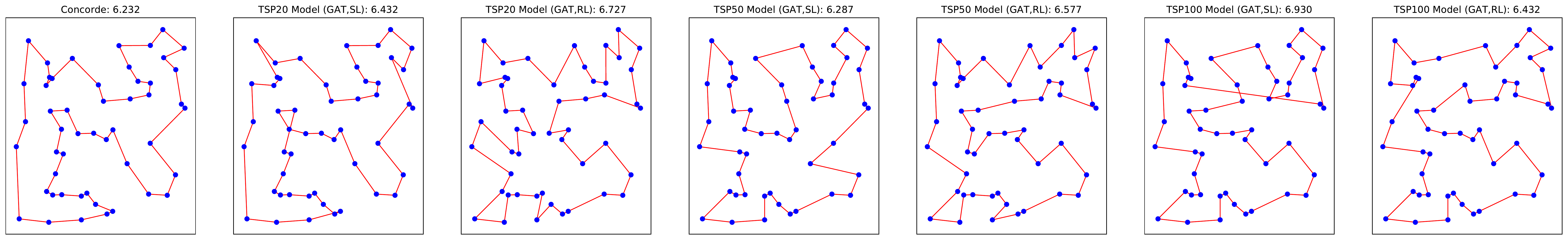}
    \label{fig:viz50-greedy}
    }
    \\
    \subfloat[TSP50 instance, Sampling (1280 solutions)]{
    \includegraphics[width=\textwidth]{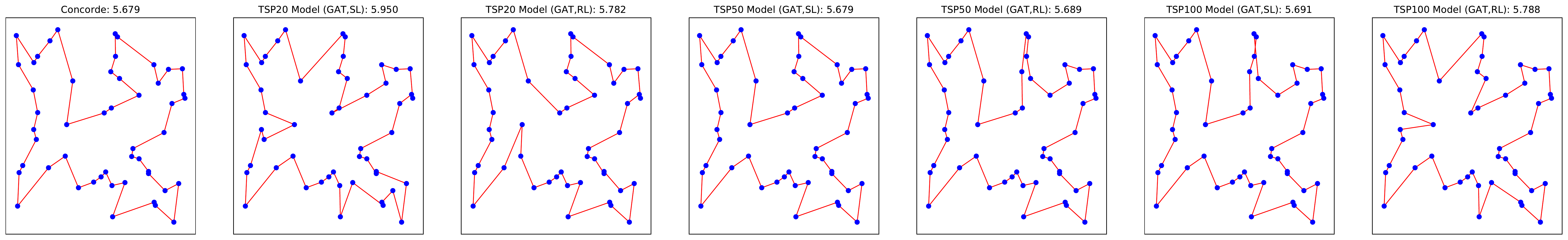}
    \label{fig:viz50-sam}
    }
    \\
    \subfloat[TSP50 instance, Beam search (1280 width)]{
    \includegraphics[width=\textwidth]{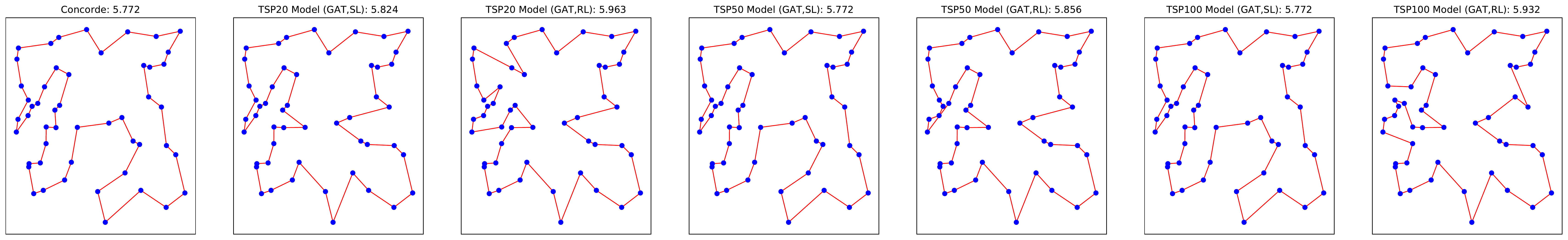}
    \label{fig:viz50-bs}
    }
    \caption{Prediction visualizations from various models for TSP50 test instances.}
    \label{fig:viz50}
\end{figure}

\begin{figure}[h!]
    \centering
    \subfloat[TSP100 instance, Greedy search]{
    \includegraphics[width=\textwidth]{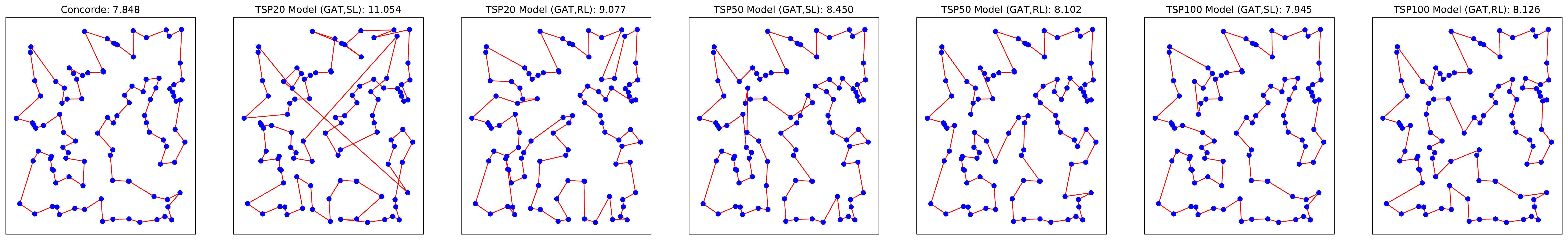}
    \label{fig:viz100-greedy}
    }
    \\
    \subfloat[TSP100 instance, Sampling (1280 solutions)]{
    \includegraphics[width=\textwidth]{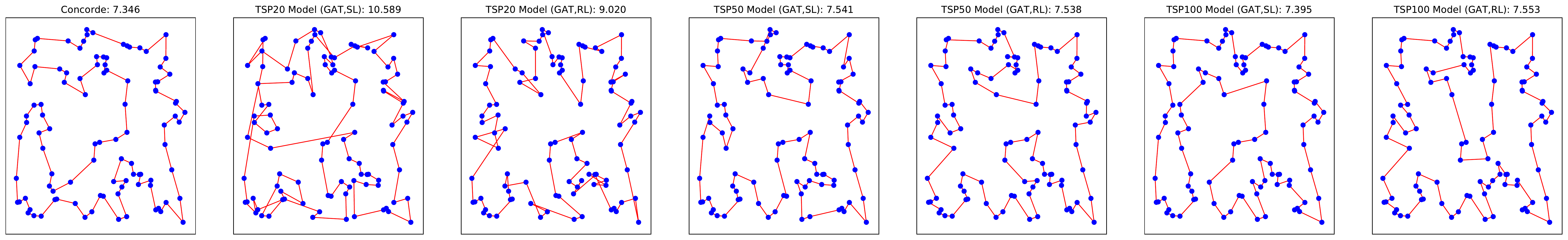}
    \label{fig:viz100-sam}
    }
    \\
    \subfloat[TSP100 instance, Beam search (1280 width)]{
    \includegraphics[width=\textwidth]{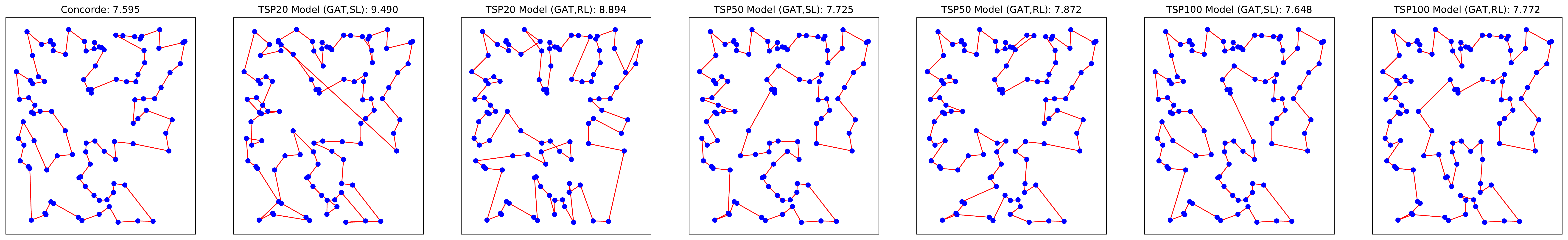}
    \label{fig:viz100-bs}
    }
    \\
    \caption{Prediction visualizations from various models for TSP100 test instances.}
    \label{fig:viz100}
\end{figure}

\begin{figure}[h!]
    \centering
    \subfloat[TSP200 instance, Greedy search]{
    \includegraphics[width=\textwidth]{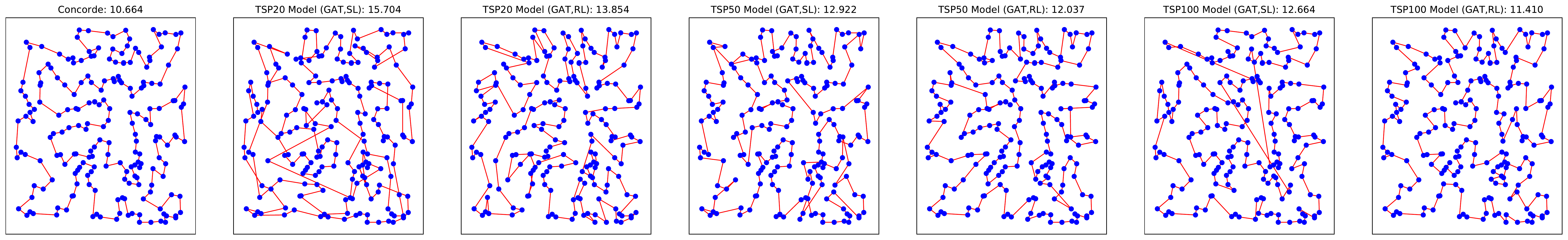}
    \label{fig:viz200-greedy}
    }
    \\
    \subfloat[TSP200 instance, Sampling (250 solutions)]{
    \includegraphics[width=\textwidth]{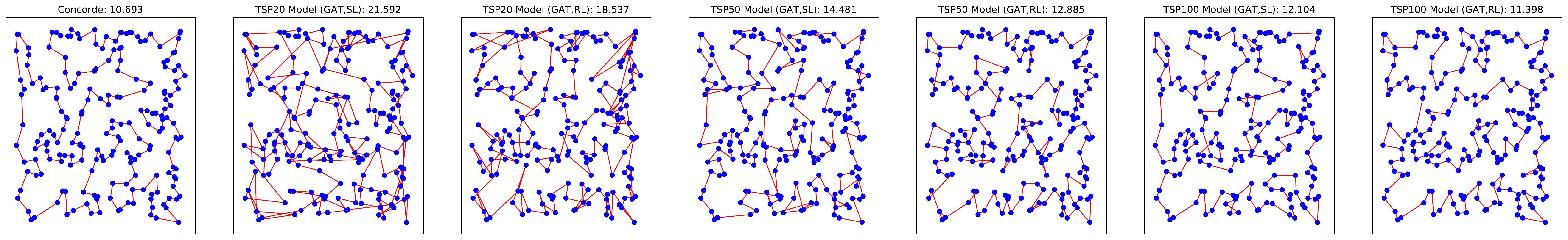}
    \label{fig:viz200-sam}
    }
    \\
    \subfloat[TSP200 instance, Beam search (250 width)]{
    \includegraphics[width=\textwidth]{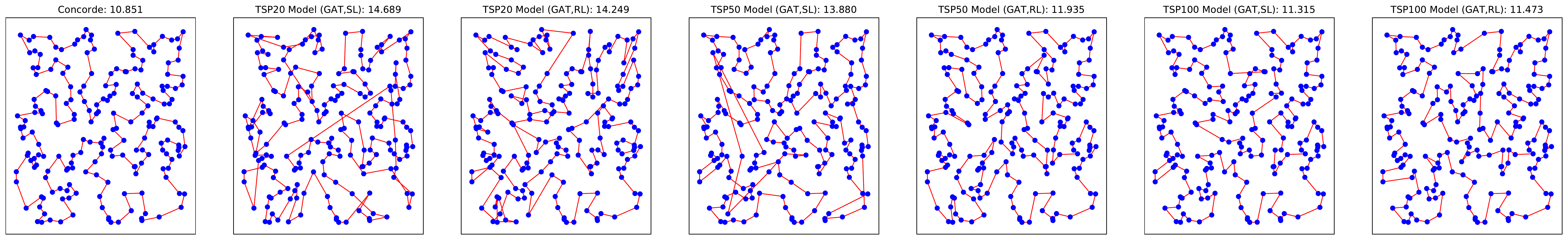}
    \label{fig:viz200-bs}
    }
    \caption{Prediction visualizations from various models for TSP200 test instances.}
    \label{fig:viz200}
\end{figure}

Figures \ref{fig:viz20}, \ref{fig:viz50}, \ref{fig:viz100} and \ref{fig:viz200} display prediction visualizations for samples from test sets of various problem
instances under various search settings. 
In each figure, the first panel shows the groundtruth
TSP tour, obtained using Concorde. 
Subsequent panels show the predicted tours from each model.

\end{document}